\documentclass[10pt]{article}
\usepackage[a4paper, top=2.5cm, bottom=2.5cm, left=2.5cm, right=2.5cm]{geometry}
\usepackage{amssymb, amsmath}
\usepackage{graphicx}
\usepackage{tabularx}
\usepackage{booktabs}
\usepackage{array}
\usepackage{algorithm}
\usepackage{algorithmicx}
\usepackage{algpseudocode}
\usepackage{listings}
\usepackage{enumitem} 
\usepackage{euscript}
\usepackage{float}
\usepackage{multirow}
\usepackage{url}
\usepackage{caption}
\usepackage{subcaption}
\usepackage{comment}
\usepackage{makecell}
\usepackage{mathrsfs}
\usepackage[numbers]{natbib}
\usepackage{soul}
\usepackage{xcolor}
\usepackage{hyperref}
\hypersetup{hidelinks}
\usepackage{amsthm}
\usepackage{fancyhdr}
\theoremstyle{definition}

\def\argmin{\mathop{\rm argmin}\limits}

\title{Designing Compact Neural Architectures via Neuron Gating and Mixed Activation}

\author{Abhishek Shukla$^{1}$ \and Ankur Sinha$^{2}$ \and Faiz Hamid$^{1}$}

\date{
$^{1}$Department of Management Sciences, IIT Kanpur, India\\
\texttt{abhiskl@iitk.ac.in, fhamid@iitk.ac.in}\\[6pt]
$^{2}$Krishnamurthy Tandon School of AI, IIM Ahmedabad, India\\
\texttt{asinha@iima.ac.in}
}

\begin{document}

\maketitle
\begin{abstract}
Neural Architecture Search (NAS) is naturally formulated as a bilevel optimization problem, where the upper-level task involves optimizing the neural architecture with respect to validation performance, while the lower-level task focuses on training the network parameters by minimizing the training loss. However, NAS, being a combinatorial optimization problem, is computationally expensive due to the discrete nature of architectural decisions, the exponential growth of the search space with network depth and width, and the high cost of training candidate architectures to optimality. This work aims to develop a fundamental method for NAS that is applicable across diverse neural network architectures---including MLPs, CNNs, RNNs, and Transformers---for identifying compact architectures with strong predictive performance. We propose three scalable and efficient bilevel optimization formulations for NAS. This replaces discrete neuron- and activation-level architectural decisions with continuous relaxations, enabling differentiable optimization over otherwise combinatorial architecture spaces. The resulting differentiable bilevel problems are efficiently solved using hypergradient-based methods, giving rise to the NAS methods termed NAS based on Neuron Gating (NAS-NG), NAS based on Mixed Activation (NAS-MA), and NAS based on Neuron Gating and Mixed Activation (NAS-NGMA). The proposed methods are evaluated on MLP and CNN architectures with varying depths and widths using the MNIST dataset. They are further assessed on the CIFAR-10 dataset against vanilla Differentiable Architecture Search (DARTS). Across these experiments, the proposed approaches consistently identify compact architectures while achieving competitive or improved predictive performance. In particular, on MNIST, NAS-NGMA achieves 98.68\% test accuracy with 7.69M MLP parameters, while NAS-NG attains 99.63\% test accuracy with only 0.26M CNN parameters. On CIFAR-10, the proposed methods consistently outperform the DARTS method. Additional experiments further show that NAS-NG scales effectively to substantially over-parameterized architectures and can further improve literature-optimal architectures through neuron-level optimization, yielding higher accuracy with significantly fewer parameters. These findings establish relaxed bilevel optimization as a practical and scalable alternative to discrete search for combinatorial NAS, one that replaces expensive architecture-level search with efficient, gradient-based optimization at the neuron and activation level. Because the formulations are agnostic to the underlying network type, they offer a general route to compact, high-performing architectures across MLPs, CNNs, and beyond, and point toward extending neuron-level optimization to further improve even well-established, hand-tuned architectures.
\end{abstract}


\section{Introduction}\label{sec1}
Artificial Neural Networks (ANNs) consist of layered, interconnected neurons whose trainable parameters (weights and biases) are optimized to minimize task-specific loss, whereas hyperparameters, such as network depth and width, activation functions, and optimization settings, are fixed prior to training. Among these, architectural parameters play a decisive role in determining a model's expressive power and generalization ability. Increasing the number of layers enhances representational capacity, but excessive capacity can result in overfitting, where the models fit the training data perfectly but generalize poorly. Traditionally, neural architectures have been designed through expert intuition and extensive trial-and-error. In the absence of domain expertise, discovering an effective architecture can be highly challenging. Furthermore, even expert-driven design does not necessarily yield optimal architectures, especially as the complexity of modern deep neural networks continues to grow. Consequently, architectural design remains a fundamental challenge in deep learning.

Neural Architecture Search (NAS), a subfield of Automated Machine Learning (AutoML) \cite{zoller2021benchmark} and Hyperparameter Optimization (HPO)  \cite{cowen2022hebo}, seeks to automate network design by systematically exploring large architecture search spaces. These efforts also connect to Automated Reinforcement Learning (AutoRL) \cite{parker2022automated}, which applies AutoML principles to RL while addressing additional challenges unique to RL agents, whose success is often highly sensitive to training design choices. The primary objective of NAS is to identify an optimal architecture from a large and complex set of design choices tailored to a specific task and dataset. A comprehensive conceptual framework for NAS is presented in \cite{elsken2019neural}. Although effective, NAS poses significant computational challenges due to the discrete nature of architectural decisions and the high cost of training many candidate architectures. This motivates optimization-based formulations that can exploit the structure in the search space. Formally, NAS, like many HPO tasks, can be naturally formulated as a Bilevel Optimization Problem (BOP)~\cite{ren2021comprehensive,shukla2026bilevel}. In this formulation, the leader's problem determines the architectural parameters, while the follower's problem optimizes the corresponding network weights through standard training.

The hierarchical separation in NAS is not merely cosmetic: it prevents architectures from being evaluated using inadequately trained weights and ensures that the search trajectory is guided by generalization rather than overfitting. In this way, bilevel optimization provides a theoretically grounded framework for NAS that disentangles architecture evaluation from weight optimization. However, the NAS search space is typically high-dimensional and discrete, rendering the optimization process both computationally demanding and algorithmically challenging. A complementary perspective on architectural discovery comes from evolutionary approaches: \citet{stanley2004competitive} demonstrate that incrementally complexifying neural architectures through evolutionary search---rather than optimizing over a fixed structure---leads to significantly more sophisticated solutions, underscoring the importance of allowing the search process itself to evolve. Despite these complexities, NAS has demonstrated remarkable success in automating architectural design, substantially reducing human intervention and producing models that match or even outperform manually crafted architectures. When another Machine Learning (ML) model---such as RL \cite{zoph2017neural}---is employed to explore the architecture space of a Neural Network (NN), the search often incurs extremely high computational costs (on the order of 2000 GPU days), primarily due to the combinatorial explosion of candidate architectures and the inherent difficulty of learning-based approaches in consistently navigating billions of possibilities to identify the best feasible (optimal) design, given the inevitable gap between predicted and optimal decisions. This limitation motivates the adoption of optimization-driven architecture search frameworks rather than relying solely on conventional ML-based or naive search strategies. In this direction, a prominent and efficient approach that avoids discrete optimization is Differentiable Architecture Search (DARTS) \cite{DBLP:conf/iclr/LiuSY19}, which relaxes the architecture search space into a continuous domain, thereby enabling a tractable bilevel optimization formulation. In recent years, there has been a significant surge in the DARTS literature, leading to the development of numerous variants of the original algorithm \cite{chen2019progressive,xu2019pcdarts,att-darts,fair-darts,zhang2023enhanced,zhu2024relax,yang2024ostr,guo2024semantic,li2024lmd,cai2024sto}. Notably, \citet{tuli2023flexibert} propose a heterogeneous and flexible transformer search space, demonstrating that relaxing homogeneity assumptions in architecture design---combined with surrogate-based bilevel optimization---yields substantially more compact and higher-performing models than their homogeneous counterparts. We also aim to obtain a high-performing and compact neural network architecture by solving a continuous bilevel NAS problem using a hypergradient-based method \cite{giovannelli2021inexact}, inspired by the DARTS framework. Beyond NAS, BOPs have also been employed for overfitting control in large language models, including transformer architectures, as demonstrated in \cite{shukla2026lift}.

\begin{figure}[ht]
    \centering
        \includegraphics[width=\linewidth]{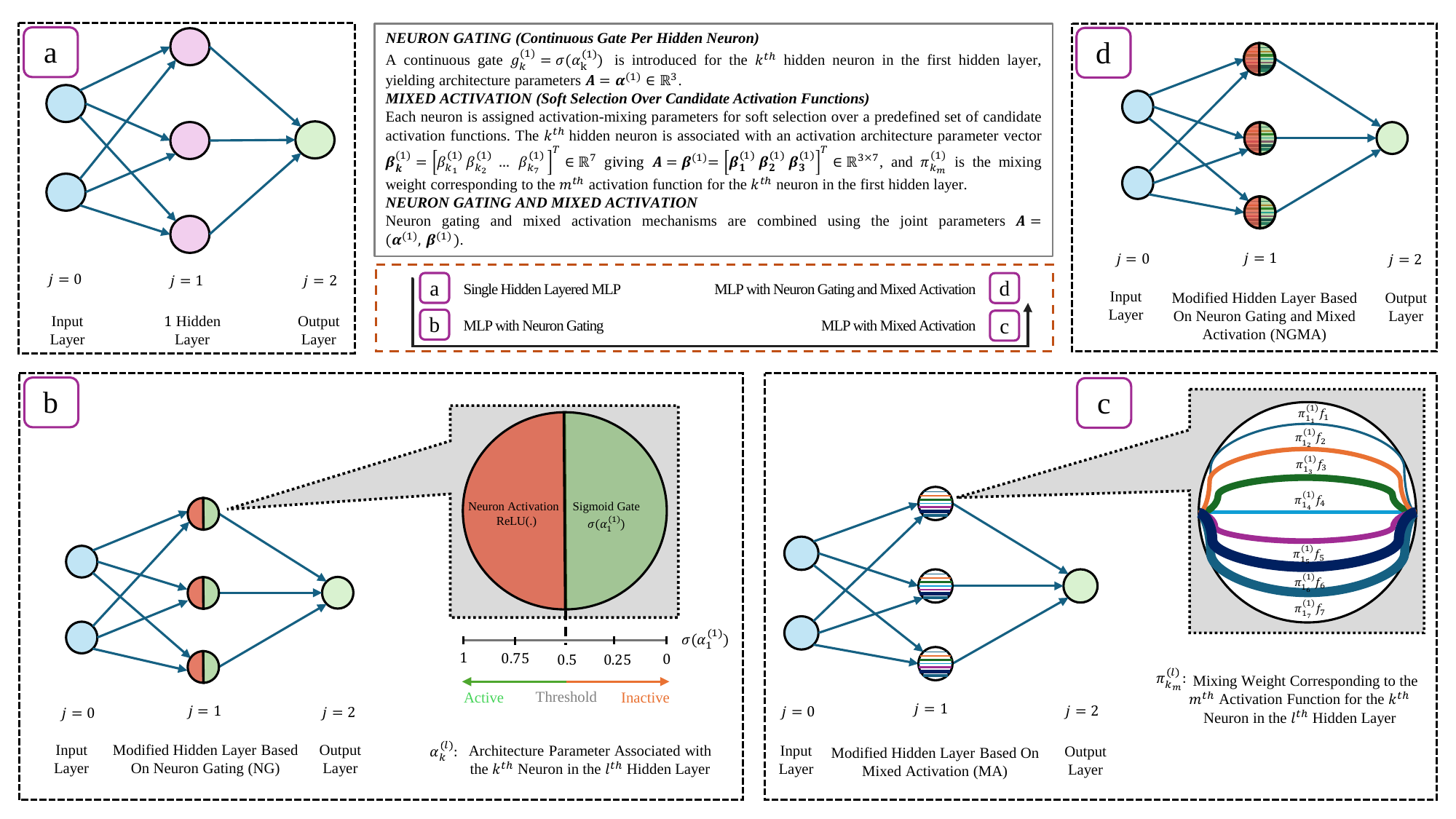}
        \caption{Progressive construction of MLP architectures via neuron- and activation-level modifications.}
        \label{mlps}
\end{figure}

This paper presents a fundamental method for NAS with the potential for broad applicability. In particular, we develop three relaxed bilevel formulations of the original combinatorial NAS problem, enabling a tractable exploration of the architecture search space. The proposed approach replaces discrete neuron- and activation-level decisions with continuous variables, resulting in differentiable bilevel programs that can be efficiently optimized. Multilayer Perceptron (MLP) and Convolutional Neural Network (CNN) architectures are employed as testbeds to systematically evaluate the effectiveness of the proposed formulations on standard benchmark datasets. Figure~\ref{mlps} illustrates the methodological progression from the original MLP architecture (a) to the proposed relaxed models (b, c, and d) based on Neuron Gating (NG), Mixed Activation (MA), and Neuron Gating and Mixed Activation (NGMA). Although the formulations are developed in the context of MLPs, the framework is general and extends naturally to CNNs and Recurrent Neural Networks (RNNs) through appropriate tensor generalizations. The effectiveness of the proposed approach is validated through a comprehensive comparative experimental evaluation of MLP and CNN architectures on the MNIST dataset and CNN architectures on the CIFAR-10 dataset. The main contributions of this work are summarized as follows:
\begin{enumerate}
    \item \textbf{Relaxed bilevel formulation:} We formulate NAS as relaxed bilevel optimization problem by replacing discrete neuron- and activation-level design choices with continuous variables, enabling differentiable and scalable optimization.
    \item \textbf{Novel NAS algorithms:} We propose three novel strategies for architecture search: NAS-NG, NAS-MA, and NAS-NGMA.
    \item \textbf{Comprehensive experimental evaluation:} We conduct a comprehensive experimental evaluation on MLP and CNN architectures for MNIST, and CNN architectures for CIFAR-10, demonstrating the effectiveness of the proposed formulations.
    \item \textbf{Superior performance over DARTS method:} We empirically show that the proposed approaches outperform vanilla DARTS on CIFAR-10.
    \item \textbf{Compact and accurate architectures:} We show that the proposed methods consistently identify compact architectures with competitive or improved predictive performance on MNIST dataset, achieving approximately 40--75\% reduction in model parameters. Similar trends are observed on the CIFAR-10.
    \item \textbf{Competitive MNIST results:} We demonstrate that the best-performing models achieve 98.68\% test accuracy with 7.69M parameters for MLPs (NAS-NGMA), and 99.63\% test accuracy with only 0.26M parameters for CNNs (NAS-NG).
    \item \textbf{Improving baseline and optimal architectures:} We show that NAS-NG remains effective when initialized from substantially over-parameterized architectures and can further improve literature-optimal architectures (architecture fine-tuning), demonstrating that neuron-level optimization complements existing topology search methods.
\end{enumerate}

The remainder of this paper is organized as follows. Section~\ref{sec2} reviews the relevant literature, and Section~\ref{sec3} presents the necessary preliminaries. Section~\ref{sec4} introduces the proposed framework. Section~\ref{sec5} presents a comprehensive empirical evaluation, including experiments on MLP and CNN architectures for the MNIST dataset, where the proposed relaxed bilevel optimization formulations are compared against baseline models. This section further analyzes optimization behavior, generalization performance, and model complexity, and highlights the key findings. In addition, the proposed NAS methods are evaluated on the CIFAR-10 dataset and compared with DARTS method. Finally, Section~\ref{sec6} concludes the paper by discussing the limitations, summarizing the main contributions, and outlining directions for future work.

\section{Related Works}\label{sec2}
In this section, we discuss advancements in neural network-based handwritten digit recognition and differentiable NAS, including classical machine learning methods, deep neural architectures, ensemble techniques, and DARTS-based NAS methods.

\subsection{Neural Architectures and Learning Methods}
Early work laid the foundation for gradient-based learning and CNNs for document recognition tasks \cite{lecun2002gradient}. Alongside deep learning approaches, classical ML methods continue to play a significant role. For example, enhancements to multi-layer logistic regression using outlier detection have been proposed in \cite{kang2018new}, while privacy-preserving learning using homomorphic encryption has been explored for logistic regression in \cite{han2019logistic}. Dimensionality reduction techniques such as Principal Component Analysis (PCA) and Linear Discriminant Analysis (LDA) have also been widely studied for handwritten digit recognition \cite{patel2019handwritten}, and advances in similarity-based methods include quantum $k$-nearest neighbor algorithms \cite{wang2019improved}. With the advent of deep learning, CNNs have consistently demonstrated superior performance over MLPs, primarily due to their ability to effectively capture spatial hierarchies in image data \cite{ben2025handwritten}. Extensive efforts have been made to improve CNN performance by analyzing architectural and training factors, including the impact of network depth and training epochs \cite{arif2018study}, as well as structural parameters such as stride and receptive fields \cite{ahlawat2020improved}. The use of deep, pre-trained architectures, such as AlexNet and GoogleNet, has further advanced performance benchmarks \cite{soomro2017performance}. Beyond conventional CNNs, biologically inspired models, such as deep spiking CNNs based on Spike Time Dependent Plasticity (STDP), have also been explored \cite{kalbande2022performance}. In addition, ensemble learning approaches that combine multiple base learners have been widely adopted to achieve state-of-the-art performance \cite{nandan2020handwritten, ullah2025handwritten}. Several high-performing models have been reported in the literature, including those in \cite{ahmed2023novel,kowsari2018rmdl,wan2013regularization,ciregan2012multi,sato2015apac}, highlighting the continued progress in this domain.\\

\subsection{Differentiable Architecture Search}
The DARTS method, introduced in \cite{DBLP:conf/iclr/LiuSY19}, presented a highly efficient and competitive NAS approach compared to RL- and Evolutionary Computation (EC)-based methods, functioning within a continuous search space. Building on DARTS, Progressive Differentiable Architecture Search (P-DARTS) was developed by \cite{chen2019progressive}, which addressed computational challenges and enhanced search stability through search space approximation and regularization techniques. Further, Robustness of Differentiable Architecture Search (R-DARTS) was explored in  \cite{zela2019understanding}  through a methodical exploration of architectural spaces and regularization strategies. In \cite{xu2019pcdarts}, the authors introduced an innovative method known as partially-connected differentiable architecture search (PC-DARTS) with the obvious goal of improving the efficiency and stability of NAS. Att-DARTS algorithm devised in \cite{att-darts} is another innovative extension of the DARTS framework, adeptly integrating attention modules into the NAS process. The manifestation of the collapse phenomenon in DARTS, characterized by an excessive occurrence of skip-connects over an extensive number of search epochs due to overfitting of the one-shot model, resulted in diminished performance. This challenge was effectively addressed by introducing early stopping criteria in DARTS+ \cite{liang2019darts+}, where DARTS stops when two or more skip-connects appear in a normal cell or when the architecture parameter ranking remains stable for a predefined number of epochs. Fair DARTS \cite{fair-darts} was introduced to tackle issues of collapse observed in the DARTS algorithm due to unfair advantages in exclusive competition among candidate operations during the search process. DE-DARTS developed by \cite{de-darts} addressed the challenges of gradient-based NAS by proposing a novel approach that incorporated Dynamic Attention Networks (DANs). More recently, EG-DARTS \cite{zhang2023enhanced} used a multi-objective evolution-based approach, combining gradient optimization with evolutionary strategies to improve DARTS’ search effectiveness. Additional advancements have been made with methods like Relax-DARTS \cite{zhu2024relax}, STO-DARTS \cite{cai2024sto}, OSTR-DARTS \cite{yang2024ostr}, semantic DARTS \cite{guo2024semantic}, LMD-DARTS \cite{li2024lmd}, HN-DARTS \cite{li2024hn}, and DARTS-EAST \cite{fang2025darts}, among others, highlighting continued innovation within the DARTS literature.

\section{Preliminaries}\label{sec3}
Formally, the bilevel NAS problem can be expressed as follows:
\begin{equation}\label{formulation_1}
	\begin{aligned}
		\min_{A} \quad & \mathcal{L}_v(A, \hat{W}) \\
		\text{s.t.} \quad & \hat{W} \in \argmin_{W \in \mathscr{W}} \mathcal{L}_t(A, W) \\
		& A \in \mathscr{A},
	\end{aligned}
\end{equation}
where \(\mathcal{L}_t\) and \(\mathcal{L}_v\) denote the training and validation losses, respectively, \(\mathscr{A}\) denotes the architecture space, and \(\mathscr{W}\) denotes the network weight space. The lower-level problem mirrors the conventional training of an NN, where the weights \(W\) are optimized to minimize the training loss for a fixed architecture \(A\). In contrast, the upper-level problem seeks architectures that generalize well, as measured by the validation loss.

The gradient of the upper-level objective (validation loss) with respect to the upper-level decision variable \(A\) is called the hypergradient. One of the most well-known optimization methods for NAS is based on an approximate hypergradient \cite{giovannelli2021inexact}. This hypergradient is used to update the architecture parameters to improve validation performance while accounting for the model weights obtained after one step of gradient descent on the training loss. The hypergradient is approximated as
\[
\nabla_A \mathcal{L}_v(A, \hat{W}) \approx \nabla_A \mathcal{L}_v\left(A, W - \xi \nabla_W \mathcal{L}_t(A, W)\right),
\]
where $\xi$ is a small learning rate used for one-step unrolled optimization of the model training problem. Setting $\xi = 0$ yields a first-order approximation, while $\xi \neq 0$ introduces a second-order correction
\begin{equation}\label{approx_hypergrad}
  \nabla_A \mathcal{L}_v(A,\hat{W}) \approx \nabla_A \mathcal{L}_v(A, W') - \xi \nabla^2_{A, W} \mathcal{L}_t(A, W) \nabla_{W'} \mathcal{L}_v(A, W'),  
\end{equation}
with $W' = W - \xi \nabla_W \mathcal{L}_t(A, W)$. The second term captures the interaction between the architecture parameters and the network weights through the mixed Hessian of the training loss, thereby accounting for their influence on the approximate hypergradient. This term can be efficiently approximated using finite-difference schemes. For further details on DARTS, see \cite{DBLP:conf/iclr/LiuSY19}.

\section{Proposed Method}\label{sec4}
\begin{figure}[ht]
\centering
	\includegraphics[width=0.8\linewidth]{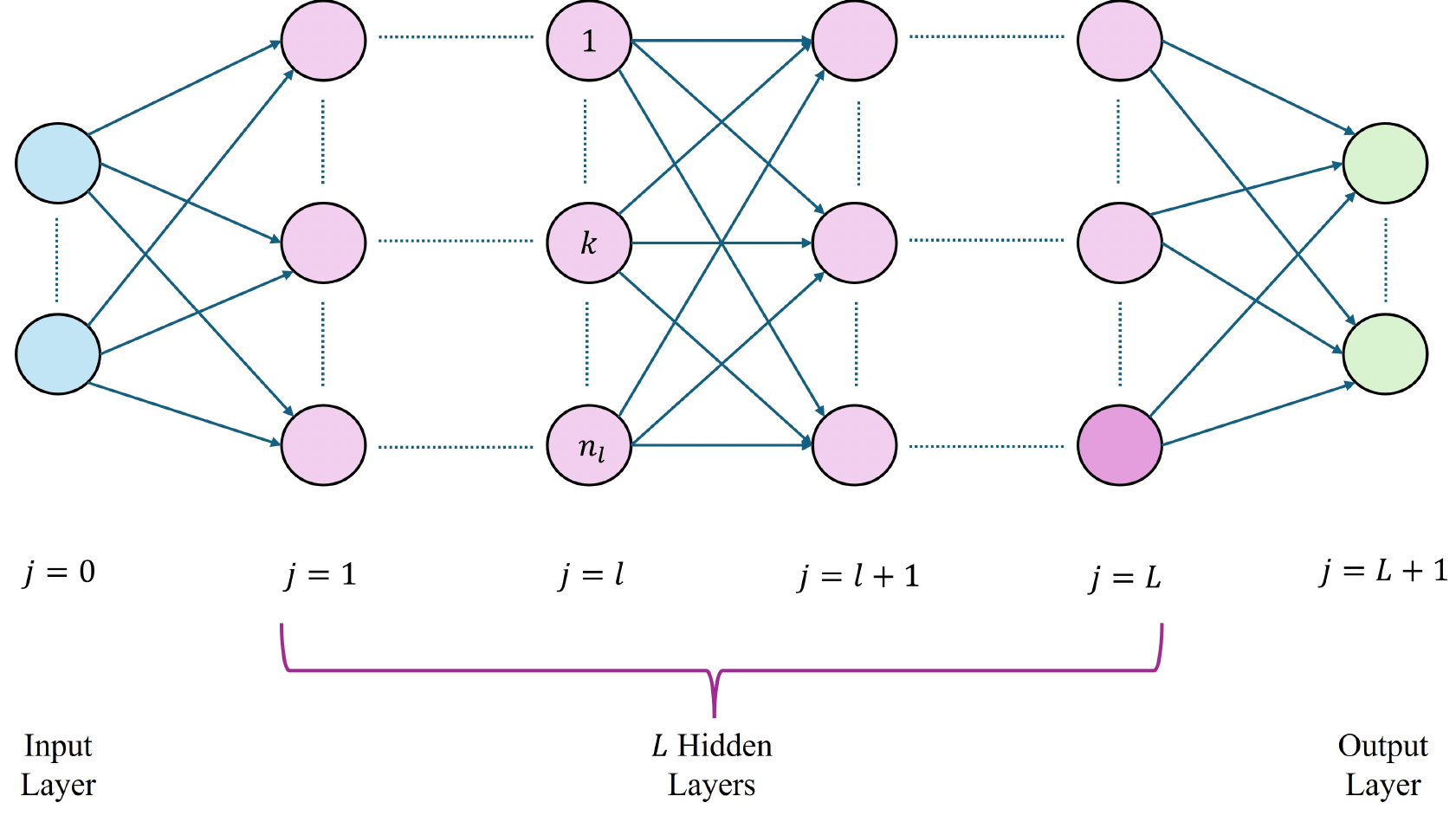}
	\caption{Schematic representation of a general MLP architecture with an input layer, $L$ hidden layers, and an output layer.}
	\label{general_mlp_arch}
\end{figure}

Consider a fully connected NN depicted in Figure~\ref{general_mlp_arch}. This network is composed of an input layer, $L$ hidden layers, and an output layer. The symbols used to represent this MLP are detailed in Table~\ref{tab:mlp_notations}.

\begin{table}[ht]
\centering
\caption{Notations used for the MLP model.}
\label{tab:mlp_notations}
\footnotesize
\renewcommand{\arraystretch}{1}
\resizebox{\linewidth}{!}{
\begin{tabular}{@{}p{4.0cm} p{9cm}@{}}
\toprule
\textbf{Notation} & \textbf{Description} \\
\midrule
$S^{T}, \, S^{V}$ & Training and validation sets \\
$\mathbf{x}_i \in \mathbb{R}^{n_0}, \,y_i \in \{0,1\}$ & Input vector and true output label for the $i^{\text{th}}$ sample \\
$j = 0$ & Index of the input layer \\
$j = 1, \dots, l,\;\; l+1, \dots, L$ & Indices of the hidden layers \\
$j = L+1$ & Index of the output layer \\
$n_j$ & Number of neurons in the $j^{\text{th}}$ layer \\
$\mathbf{W}^{(j)} \in \mathbb{R}^{n_{j+1} \times n_j}$ & Weight matrix connecting layer $j$ to layer $j+1$ \\
$\mathbf{b}^{(j)} \in \mathbb{R}^{n_{j+1}}$ & Bias vector associated with the $(j+1)^{\text{th}}$ layer \\
$f(\cdot)$ & Element-wise activation function (ReLU) \\
$\mathbf{z}_i^{(j)}$ & Pre-activation vector of the $(j+1)^{\text{th}}$ layer for the $i^{\text{th}}$ sample \\
$\mathbf{h}_i^{(j)} = f(\mathbf{z}_i^{(j)})$ & Activation vector of the $j^{\text{th}}$ layer for the $i^{\text{th}}$ sample \\
$\mathcal{J} = \{0, 1, \dots, L+1\}$ & Set of all layer indices \\
$\mathcal{K}_j = \{1, \dots, n_j\}$ & Set of neuron indices in the $j^{\text{th}}$ layer \\
\bottomrule
\end{tabular}
}
\end{table}

\begin{figure}[ht]
\centering
	\includegraphics[width=0.375\linewidth]{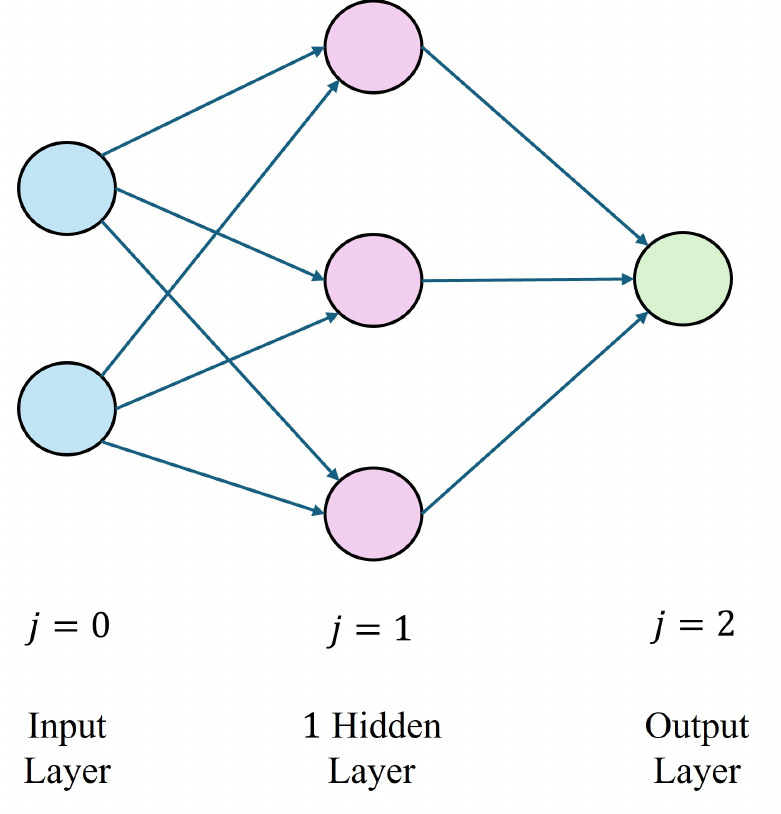}
	\caption{Neural network with one hidden layer.}
	\label{single_layer_mlp}
\end{figure}

The structure of an MLP can be defined by the count of neurons present in its input, hidden, and output layers, and is depicted as a tuple
\[
(\text{Input} = [n_0],\; \text{Hidden} = [n_1, \dots, n_l,\;\; n_{l+1}, \dots, n_L],\; \text{Output} = [n_{L+1}])
\]
This representation allows for heterogeneous hidden layers with varying widths. To illustrate, imagine a feedforward NN that has two neurons for input, a single hidden layer with three neurons, and one neuron for output., i.e.,
\[
\text{Architecture} = (\text{Input} = [2],\; \text{Hidden} = [3],\; \text{Output} = [1]),
\]
as shown in Figure~\ref{single_layer_mlp}. This network is designed for binary classification. The hidden layer's neurons utilize the ReLU activation function, whereas the output neuron employs a sigmoid activation function to generate a probabilistic output that corresponds to the input vector $\mathbf{x}_i$ given as follows:
\begin{equation}
\mathbf{x}_i =
\begin{bmatrix}
x_{1i} \\
x_{2i}
\end{bmatrix}
\in \mathbb{R}^{2}
\end{equation}

The hidden layer is parameterized by a weight matrix $\mathbf{W}^{(1)} \in \mathbb{R}^{3 \times 2}$ and a bias vector $\mathbf{b}^{(1)} \in \mathbb{R}^{3}$, while the output layer is parameterized by a weight vector $\mathbf{W}^{(2)} \in \mathbb{R}^{1 \times 3}$ and a scalar bias $b^{(2)} \in \mathbb{R}$. Let the collection of all network weights and biases be denoted by $\mathbf{W}
=
\{\mathbf{W}^{(1)}, \mathbf{b}^{(1)}, \mathbf{W}^{(2)}, b^{(2)}\}
\in \mathbb{R}^{13}$. We can write the binary cross-entropy loss as follows (derived in Appendix~\ref{loss_fn}).
\begin{equation}
\begin{aligned}
\mathcal{L}_i(\mathbf{W}; \mathbf{x}_i, y_i)
=
- \Bigg[
& y_i
\log\!\left(
\frac{1}{
1 + \exp\!\left(
-\left[
\sum_{k=1}^{3}
w^{(2)}_{k}
\max\!\left(
0,\,
w^{(1)}_{k1} x_{1i}
+
w^{(1)}_{k2} x_{2i}
+
b^{(1)}_{k}
\right)
+ b^{(2)}
\right]
\right)
}
\right)
\\[0.5em]
& \hspace{-2.5cm} 
+ (1 - y_i)
\log\!\left(
1 -
\frac{1}{
1 + \exp\!\left(
-\left[
\sum_{k=1}^{3}
w^{(2)}_{k}
\max\!\left(
0,\,
w^{(1)}_{k1} x_{1i}
+
w^{(1)}_{k2} x_{2i}
+
b^{(1)}_{k}
\right)
+ b^{(2)}
\right]
\right)
}
\right)
\Bigg]
\end{aligned}
\end{equation}\\
\noindent \textbf{Model Training Problem.}\\
The optimal model parameters are obtained by solving an unconstrained, highly nonlinear optimization problem defined by minimizing the loss function over the model’s parameter space for all the training samples, as given below:
\begin{equation}\label{model_training}
\begin{aligned}
\min_{\mathbf{W} \in \mathbb{R}^{13}} \quad
& \mathcal{L}_{t}\!\left(\mathbf{W}\right)
=
\frac{1}{|S^{T}|}
\sum_{(\mathbf{x}_i, y_i) \in S^{T}}
\mathcal{L}_i\!\left(\mathbf{W};\, \mathbf{x}_i, y_i\right)
\end{aligned}
\end{equation}
Considering a general model parameter space $\mathscr{W}$, this problem can be generalized as
\begin{equation}\label{general_model_training}
\begin{aligned}
\min_{\mathbf{W} \in \mathscr{W}} \quad
& \mathcal{L}_{t}\!\left(\mathbf{W}\right)
=
\frac{1}{|S^{T}|}
\sum_{(\mathbf{x}_i, y_i) \in S^{T}}
\mathcal{L}_i\!\left(\mathbf{W};\, \mathbf{x}_i, y_i\right)
\end{aligned}
\end{equation}
The training problem is commonly solved using gradient-based optimization techniques, including Adaptive Moment Estimation (Adam), Stochastic Gradient Descent (SGD), and several other related first-order methods.

\subsection{Bilevel Formulations for MLP Architecture Search}
In this section, we develop bilevel formulations for MLP architecture search under three settings. First, we consider neuron gating, where architecture parameters (associated with hidden neurons) control the effective participation of individual hidden neurons. Second, we investigate mixed activation function-based architecture search, in which each hidden neuron is associated with a mixed activation function (a convex combination of activation functions from a predefined set of candidate operations). By optimizing the corresponding mixing coefficients, the search process ultimately selects a single optimal activation function for each neuron. In the third setting, we unify both perspectives by jointly optimizing neuron selection and activation function within a single bilevel framework. We next describe these three settings as follows.
\subsubsection{Neuron Gating (NG)-based MLP Architecture Search}
Consider the following architecture parameter vector defined for neuron gating in the $l^{\text{th}}$ hidden layer
\begin{equation}
\boldsymbol{\alpha}^{(l)}
=
\begin{bmatrix}
\alpha^{(l)}_{1} &
\alpha^{(l)}_{2} &
\cdots &
\alpha^{(l)}_{n_l}
\end{bmatrix}^{\top}
\in \mathbb{R}^{n_l},
\qquad l = 1, \dots, L
\end{equation}
Here, $\alpha^{(l)}_{k}$ denotes the architecture parameter associated with the $k^{\text{th}}$ neuron in the $l^{\text{th}}$ hidden layer. Each $\alpha^{(l)}_{k}$ is an unconstrained real-valued variable, while the corresponding effective neuron activation gate is obtained via the sigmoid mapping
\[
\sigma\big(\alpha^{(l)}_{k}\big) \in (0,1)
\]
The index $k$ ranges from $1$ to $n_l$, and the index $l = 1, \dots, L$ enumerates the hidden layers. The global architecture parameter vector across all hidden layers is then defined as
\begin{equation}
\mathbf{A}
=
\big(
\boldsymbol{\alpha}^{(1)},
\boldsymbol{\alpha}^{(2)},
\dots,
\boldsymbol{\alpha}^{(L)}
\big)
\end{equation}

\begin{figure}[htpb]
\centering
	\includegraphics[width=\linewidth]{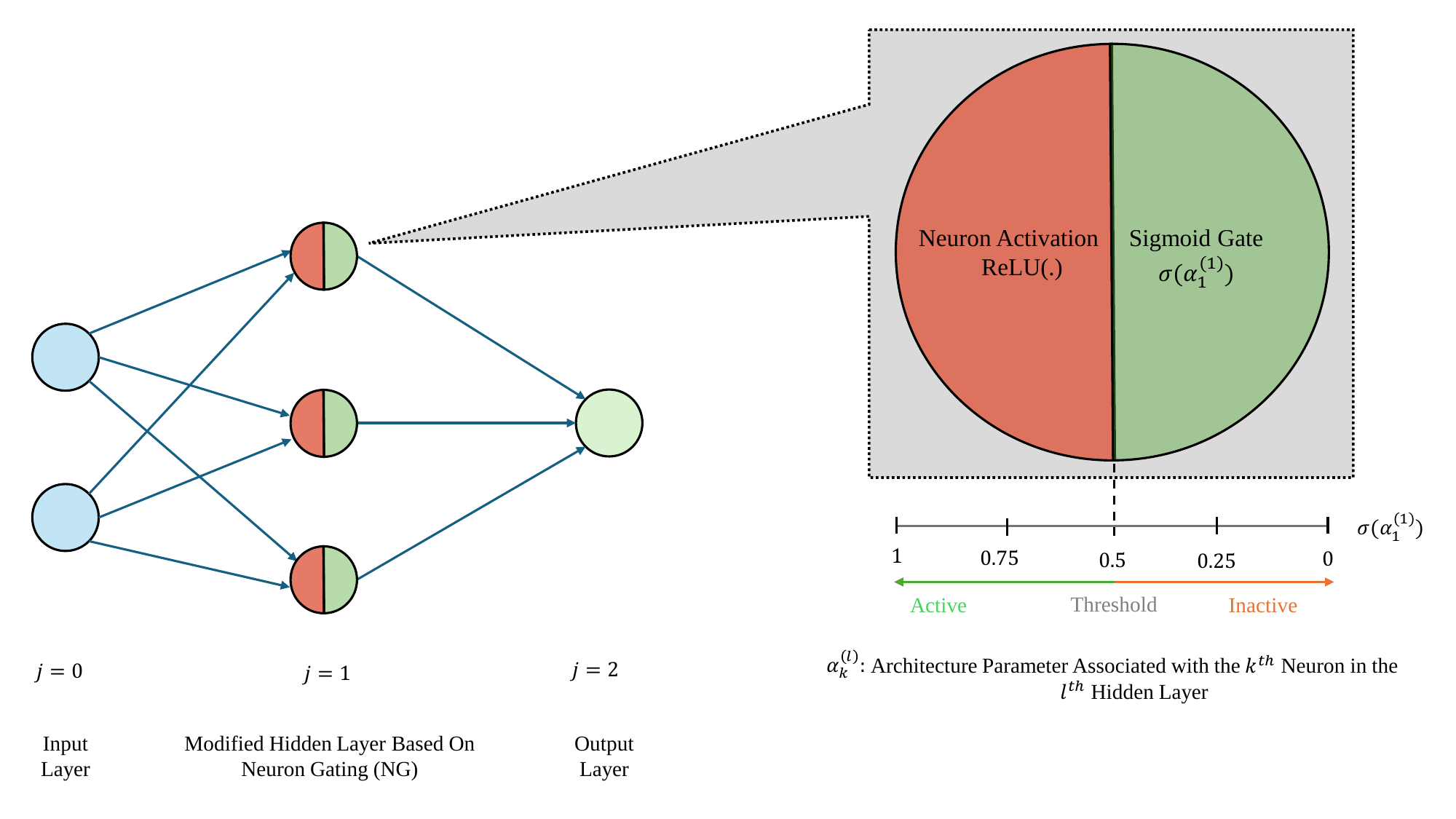}
	\caption{Neural network architecture illustrating the neuron gating mechanism employed for MLP architecture search.}
	\label{single_layer_mlp_1}
\end{figure}
\noindent For the simple modified neural network shown in Figure~\ref{single_layer_mlp_1}, the global architecture parameter vector is given by
\begin{equation}
\mathbf{A}
=
\begin{bmatrix}
\alpha^{(1)}_{1} &
\alpha^{(1)}_{2} &
\alpha^{(1)}_{3}
\end{bmatrix}^{\top}
\in \mathbb{R}^{3}
\end{equation}
The modified binary cross-entropy loss function used for architecture search is given as follows (derived in Appendix~\ref{mod_loss_fn}):
\begin{equation}\label{arch_search_loss}
\begin{aligned}
\mathcal{L}_i(\mathbf{A}, \mathbf{W}; \mathbf{x}_i, y_i)=- \Bigg[
& y_i
\log\!\left(
\frac{1}{
1 + \exp\!\left(
-\left[
\sum_{k=1}^{3}
w^{(2)}_{k}
\frac{1}{1 + \exp\!\left(-\alpha^{(1)}_{k}\right)}
\max\!\left(
0,\,
w^{(1)}_{k1} x_{1i}
+
w^{(1)}_{k2} x_{2i}
+
b^{(1)}_{k}
\right)
+ b^{(2)}
\right]
\right)
}
\right)
\\[0.6em]
& \hspace{-2.5cm}
+ (1 - y_i)
\log\!\left(
1 -
\frac{1}{
1 + \exp\!\left(
-\left[
\sum_{k=1}^{3}
w^{(2)}_{k}
\frac{1}{1 + \exp\!\left(-\alpha^{(1)}_{k}\right)}
\max\!\left(
0,\,
w^{(1)}_{k1} x_{1i}
+
w^{(1)}_{k2} x_{2i}
+
b^{(1)}_{k}
\right)
+ b^{(2)}
\right]
\right)
}
\right)
\Bigg]
\end{aligned}
\end{equation}

\noindent \textbf{MLP Architecture Search Problem Formulation.}\\
The resulting MLP architecture search problem is formulated as the bilevel optimization problem below, where $\mathcal{L}_i\!\left(.\right)$ is given by \eqref{arch_search_loss}:

\begin{equation}\label{formulation_mlp_nas_1}
\begin{aligned}
\min_{\mathbf{A} \in \mathbb{R}^{3}} \quad
& \mathcal{L}_{v}\!\left(\mathbf{A}, \hat{\mathbf{W}}\right)
=
\frac{1}{|S^{V}|}
\sum_{(\mathbf{x}_i, y_i) \in S^{V}}
\mathcal{L}_i\!\left(\mathbf{A}, \hat{\mathbf{W}};\, \mathbf{x}_i, y_i\right)
\\[0.6em]
\text{s.t.} \quad
& \hat{\mathbf{W}}
\in
\argmin_{\mathbf{W} \in \mathbb{R}^{13}}
\mathcal{L}_{t}\!\left(\mathbf{A}, \mathbf{W}\right)
=
\frac{1}{|S^{T}|}
\sum_{(\mathbf{x}_i, y_i) \in S^{T}}
\mathcal{L}_i\!\left(\mathbf{A}, \mathbf{W};\, \mathbf{x}_i, y_i\right)
\end{aligned}
\end{equation}
Considering general model parameter space ($\mathscr{W}$) and architecture parameter sapce ($\mathscr{A}$), the architecture search problem can be generalized as
\begin{equation}\label{general_architecture_search_problem}
\begin{aligned}
\min_{\mathbf{A} \in \mathscr{A}} \quad
& \mathcal{L}_{v}\!\left(\mathbf{A}, \hat{\mathbf{W}}\right)
=
\frac{1}{|S^{V}|}
\sum_{(\mathbf{x}_i, y_i) \in S^{V}}
\mathcal{L}_i\!\left(\mathbf{A}, \hat{\mathbf{W}};\, \mathbf{x}_i, y_i\right)
\\[0.6em]
\text{s.t.} \quad
& \hat{\mathbf{W}}
\in
\argmin_{\mathbf{W} \in \mathscr{W}}
\mathcal{L}_{t}\!\left(\mathbf{A}, \mathbf{W}\right)
=
\frac{1}{|S^{T}|}
\sum_{(\mathbf{x}_i, y_i) \in S^{T}}
\mathcal{L}_i\!\left(\mathbf{A}, \mathbf{W};\, \mathbf{x}_i, y_i\right)
\end{aligned}
\end{equation}

\subsubsection{Mixed Activation (MA)-based MLP Architecture Search}
In this case, the architecture search is performed by learning the activation function associated with each hidden neuron. Let $\mathcal{F} = \{f_1, f_2, \dots, f_M\}$ denote a predefined set of $M$ candidate activation functions, respectively. Each hidden neuron $k$ in the $l^{\text{th}}$ hidden layer is associated with an activation architecture parameter vector given by
\begin{equation}
\boldsymbol{\beta}^{(l)}_{k}
=
\begin{bmatrix}
\beta^{(l)}_{k_1} &
\beta^{(l)}_{k_2} &
\cdots &
\beta^{(l)}_{k_M}
\end{bmatrix}^{\top}
\in \mathbb{R}^{M}
\end{equation}
Collecting the activation architecture parameters for all neurons in the $l^{\text{th}}$ hidden layer, we define
\begin{equation}
\boldsymbol{\beta}^{(l)}
=
\begin{bmatrix}
\boldsymbol{\beta}^{(l)}_{1} &
\boldsymbol{\beta}^{(l)}_{2} &
\cdots &
\boldsymbol{\beta}^{(l)}_{n_l}
\end{bmatrix}^{\top}
\in \mathbb{R}^{n_l \times M},
\qquad l = 1, \dots, L
\end{equation}
The global activation architecture parameter across all hidden layers is then defined as
\begin{equation}
\mathbf{A}
=
\big(
\boldsymbol{\beta}^{(1)},
\boldsymbol{\beta}^{(2)},
\dots,
\boldsymbol{\beta}^{(L)}
\big)
\end{equation}
For a given hidden neuron $k$ in layer $l$, the mixing weight corresponding to the $m^{\text{th}}$ activation function is obtained via a softmax mapping given as follows
\begin{equation}
\pi^{(l)}_{k_m}
=
\frac{\exp(\beta^{(l)}_{k_m})}
{\sum_{m'=1}^{M} \exp(\beta^{(l)}_{k_{m'}})},
\quad m = 1, \dots, M
\end{equation}
\begin{figure}
\centering
	\includegraphics[width=\linewidth]{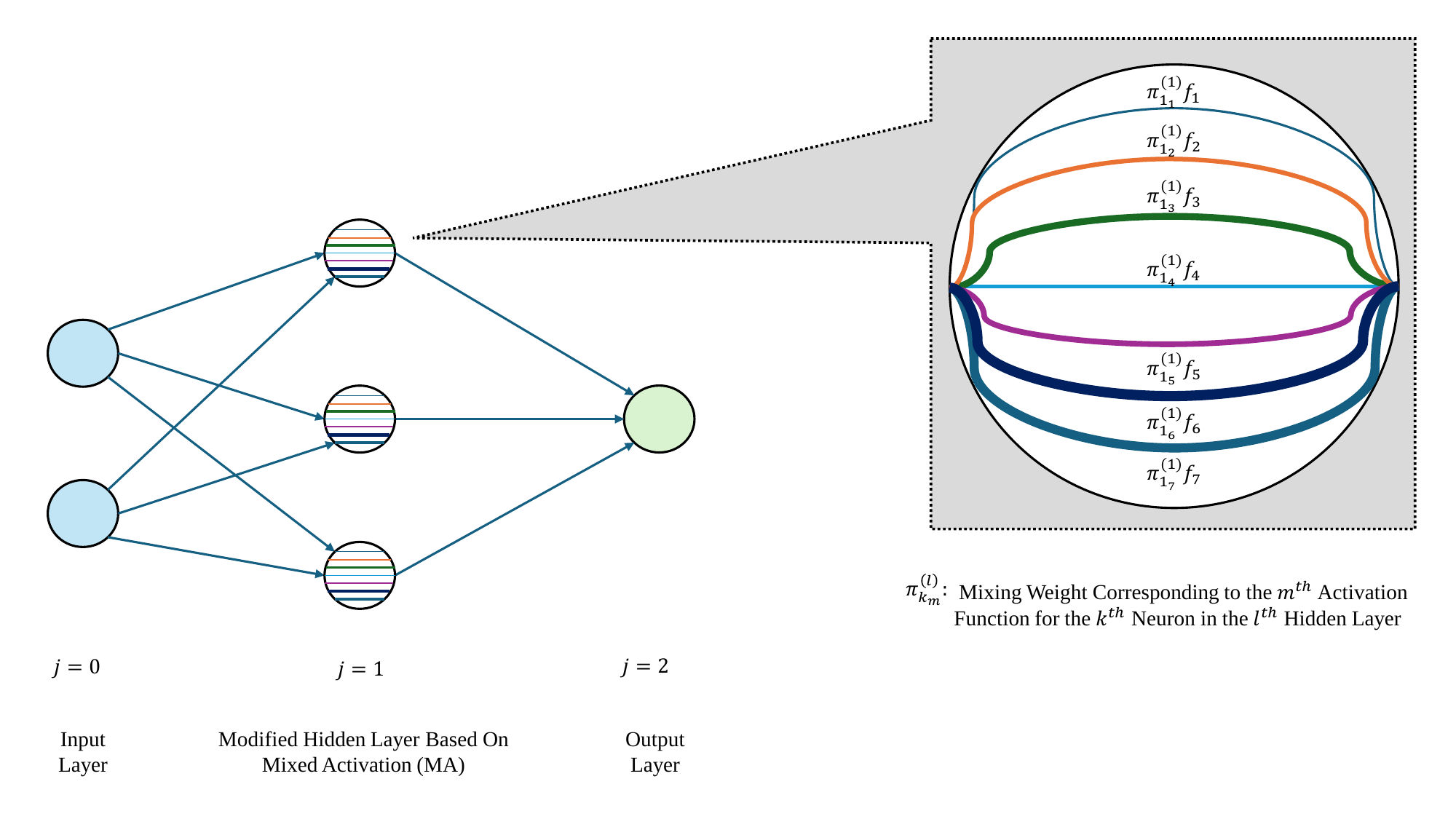}
	\caption{Modified neural network architecture for mixed activation-based architecture search.}
	\label{single_layer_mlp_2}
\end{figure}

\noindent For the simple modified neural network shown in Figure~\ref{single_layer_mlp_2}, we consider \( M = 7 \) candidate activation functions, $\mathcal{F} = \{f_1, f_2, f_3, f_4, f_5, f_6, f_7\} = \{\text{None}, \text{Identity}, \text{ReLU}, \text{Leaky ReLU}, \tanh, \sigma, \text{SiLU}\}$, for each hidden neuron, as given in Appendix~\ref{act_fns}. Since the network contains a single hidden layer with three neurons, the global architecture parameter matrix is given by
\begin{equation}\label{arch_param_act}
\mathbf{A}
=
\boldsymbol{\beta}^{(1)}
\in \mathbb{R}^{3 \times 7}
\end{equation}
Accordingly, the mixing weights for the activation functions of each hidden neuron are computed as
\begin{equation}
\pi^{(1)}_{k_m}
=
\frac{\exp(\beta^{(1)}_{k_m})}
{\sum_{m'=1}^{7} \exp(\beta^{(1)}_{k_{m'}})},
\quad m = 1, \dots, 7
\end{equation}
The activation of the $k^{\text{th}}$ hidden neuron for the $i^{\text{th}}$ input sample is then defined as a convex combination of the candidate activation functions
\begin{equation}
h'^{(1)}_{ki}
=
\sum_{m=1}^{7}
\pi^{(1)}_{k_m}
f_m\!\left(
w^{(1)}_{k1} x_{1i}
+
w^{(1)}_{k2} x_{2i}
+
b^{(1)}_{k}
\right)
\end{equation}
The pre-activation of the output neuron is given by
\begin{equation}
z_i'^{(2)}
=
\sum_{k=1}^{3}
w^{(2)}_{k} h'^{(1)}_{ki}
+ b^{(2)},
\end{equation}
and the predicted output is obtained via a sigmoid activation
\begin{equation}
\hat{y}_i' = \sigma\!\left(z_i'^{(2)}\right)
\end{equation}
Given the ground-truth label $y_i \in \{0,1\}$, the modified binary cross-entropy loss for the $i^{\text{th}}$ sample is defined as
\begin{equation}
\mathcal{L}_i(\mathbf{A}, \mathbf{W}; \mathbf{x}_i, y_i)
=
- \left[
y_i \log(\hat{y}_i')
+
(1 - y_i)\log(1 - \hat{y}_i')
\right]
\end{equation}\\
\textbf{MLP Architecture Search Problem Formulation.}\\
The MA-based MLP architecture search problem is formulated as in~\eqref{formulation_mlp_nas_1}, with the architecture parameters specified by~\eqref{arch_param_act}.

\subsubsection{Neuron Gating and Mixed Activation (NGMA)-based MLP Architecture Search}
\begin{figure}
\centering
	\includegraphics[width=0.6\linewidth]{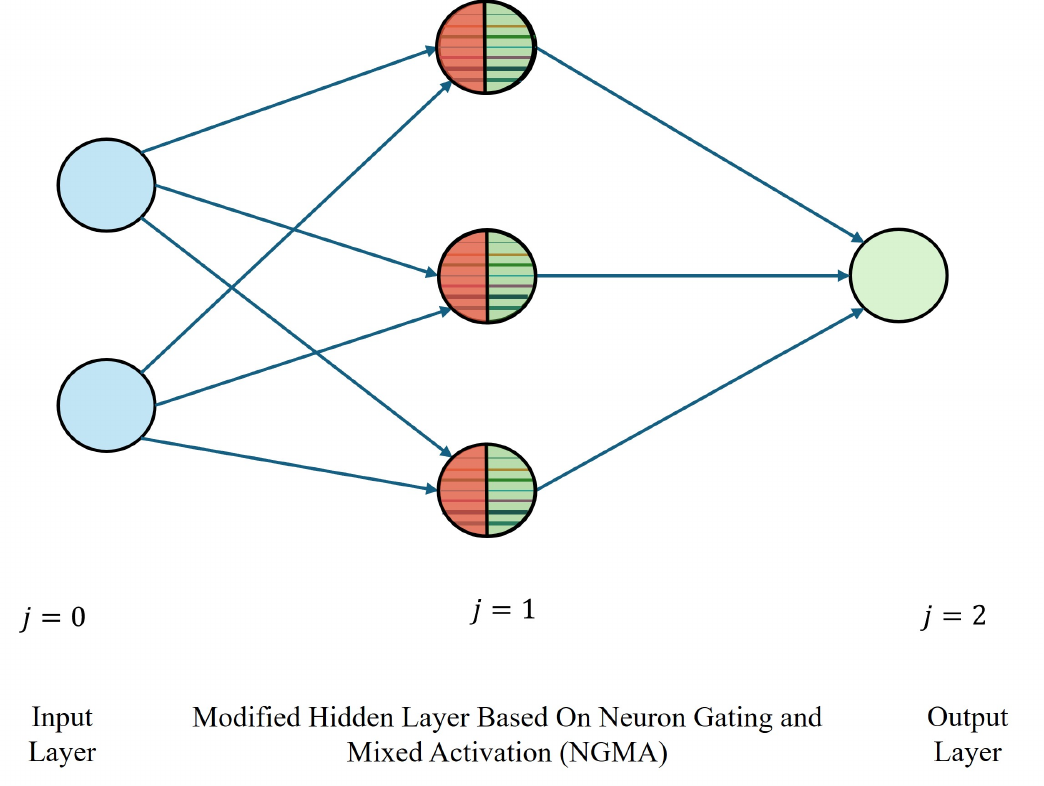}
	\caption{Neural network incorporating neuron gating mechanism and mixed activation function for MLP architecture search.}
	\label{single_layer_mlp_3}
\end{figure}

In this case, neuron selection and activation-function optimization are performed jointly within a unified architecture search framework. Each hidden neuron $k$ in the $l^{\text{th}}$ hidden layer is associated with two sets of architecture parameters
\begin{enumerate}[label=(\alph*)]
    \item a neuron gating parameter $\alpha^{(l)}_{k} \in \mathbb{R}$, and
    \item an activation architecture parameter vector $\boldsymbol{\beta}^{(l)}_{k} \in \mathbb{R}^{M}$
\end{enumerate}
The neuron gating corresponding to neuron $k$ in layer $l$ is defined via a sigmoid mapping as
\begin{equation}
g^{(l)}_{k} = \sigma\!\left(\alpha^{(l)}_{k}\right),
\end{equation}
which controls the neuron's effective contribution to the network output. The mixing weights corresponding to the candidate activation functions are obtained using a softmax mapping
\begin{equation}
\pi^{(l)}_{k_m}
=
\frac{\exp(\beta^{(l)}_{k_m})}
{\sum_{m'=1}^{M} \exp(\beta^{(l)}_{k_{m'}})},
\quad m = 1, \dots, M
\end{equation}
\noindent For the simple modified MLP shown in Figure~\ref{single_layer_mlp_3}, which consists of a single hidden layer (\(l = 1\)) and seven candidate activation functions (\(M = 7\)), the neuron-gating and activation-mixing weights reduce to
\begin{equation}
g^{(1)}_{k} = \sigma\!\left(\alpha^{(1)}_{k}\right),
\end{equation}
and
\begin{equation}
\pi^{(1)}_{k_m}
=
\frac{\exp(\beta^{(1)}_{k_m})}
{\sum_{m'=1}^{7} \exp(\beta^{(1)}_{k_{m'}})},
\quad m = 1, \dots, 7
\end{equation}
The activation of the $k^{\text{th}}$ hidden neuron for the $i^{\text{th}}$ input sample is then defined as
\begin{equation}
h'^{(1)}_{ki}
=
g^{(1)}_{k}
\sum_{m=1}^{7}
\pi^{(1)}_{k_m}
f_m\!\left(
w^{(1)}_{k1} x_{1i}
+
w^{(1)}_{k2} x_{2i}
+
b^{(1)}_{k}
\right)
\end{equation}
The output pre-activation, predicted output, and the corresponding loss function follow analogously to the previous cases. The global architecture parameter vector for the joint neuron gating and activation-based architecture search is given by
\begin{equation}\label{arch_param_switch_act}
\mathbf{A}
=
\big(
\boldsymbol{\alpha}^{(1)},
\boldsymbol{\beta}^{(1)}
\big),
\end{equation}
where $\boldsymbol{\alpha}^{(1)}$ and $\boldsymbol{\beta}^{(1)}$ denote the neuron gating and mixed activation architecture parameters of the hidden layer, respectively.\\

\noindent \textbf{MLP Architecture Search Problem Formulation.}\\
The NGMA-based MLP architecture search problem is formulated as in~\eqref{formulation_mlp_nas_1}, with the architecture parameters specified by~\eqref{arch_param_switch_act}.

\subsection{Algorithm}
\begin{figure}[ht]
\centering
	\includegraphics[width=\linewidth]{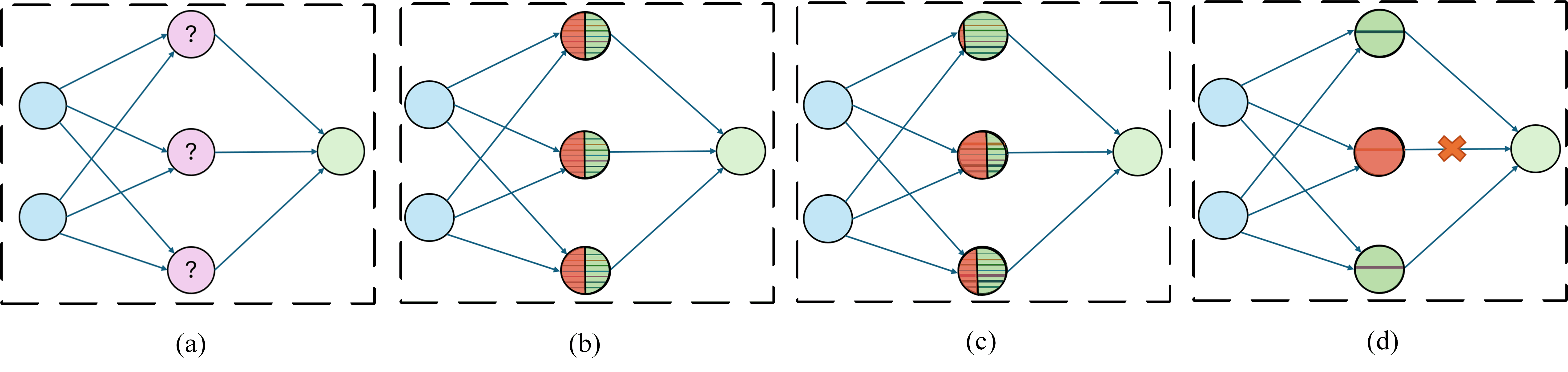}
    \caption{An overview of NAS-NGMA. (a) The gating states and activation functions of neurons are initially unknown. (b) Continuous relaxation of the search space through neuron gating and a weighted mixture of candidate activation functions for each neuron. (c) Joint optimization of the gating parameters, mixing probabilities, and network weights via a bilevel optimization framework. (d) Derivation of the final neural architecture from the learned gating states and activation-function mixing probabilities.}
	\label{neuron_gating_activation}
\end{figure}

Algorithm~\ref{mlp_neuron_gating_activation} details the proposed architecture search procedure based on NGMA, while Figure~\ref{neuron_gating_activation} provides a visual illustration.
\begin{algorithm}
\caption{NAS-NGMA}
\label{mlp_neuron_gating_activation}
\begin{algorithmic}[1]
\setlength{\itemsep}{3pt}
\State Initialize architecture parameters $A^{0}$ and network weights $W^{0}$, and set iteration counter $k \gets 0$
\While{$k < k^{\max}$}
    \State Approximate the hypergradient using equation \eqref{approx_hypergrad}
    \[
    g_A^{k} =
    \nabla_A \mathcal{L}_v
    \Bigl(
        A^{k},\;
        W^{k} - \xi \nabla_W \mathcal{L}_t(A^{k}, W^{k})
    \Bigr)
    \]
    \State Update architecture parameters
    \[
    A^{k+1} \gets A^{k} - \eta_A \, g_A^{k}
    \]
    \State Update network weights
    \[
    W^{k+1} \gets W^{k} - \eta_W \,
    \nabla_W \mathcal{L}_t(A^{k+1}, W^{k})
    \]
    \State Increment iteration counter $k \gets k+1$
\EndWhile
\State Derive the final neuron configuration by thresholding the sigmoid-transformed gating components of $A^{k^{\max}}$ and assigning to each retained neuron the activation function with the largest activation-related component in $A^{k^{\max}}$
\end{algorithmic}
\end{algorithm}
The workflow illustrated in Figure~\ref{neuron_gating_activation} consists of four main stages
\begin{enumerate}[label=(\alph*)]
\item \textbf{Problem formulation.} A sufficiently large baseline network with predetermined depth and width is constructed. At this stage, the neuron gating states and activation functions are not fixed; instead, they are treated as architecture variables to be learned through the bilevel optimization search process. For illustration, Figure~\ref{neuron_gating_activation} (a) considers a simple network with a single hidden layer containing three neurons whose gating states and activation functions are initially unknown.
\item \textbf{Continuous relaxation.} To enable gradient-based optimization, the discrete architecture decisions are transformed into continuous variables. Specifically, neuron selection is modeled through a gating mechanism, while each neuron is associated with a weighted mixture of candidate activation functions, as shown in Figure~\ref{neuron_gating_activation} (b).
\item \textbf{Optimization.} The network weights, neuron-gating parameters, and activation-function mixing probabilities are optimized jointly within a bilevel optimization framework. During this stage, the gating parameters and mixing probabilities evolve continuously, allowing the search process to identify the relative importance of individual neurons and activation functions. Figure~\ref{neuron_gating_activation} (c) depicts an intermediate optimization state.
\item \textbf{Architecture derivation.} After optimization, the learned continuous variables are converted into a discrete architecture. Neurons are retained or pruned according to the learned gating parameters and the prescribed pruning criterion. For each retained neuron, the activation function corresponding to the largest mixing probability is selected. The resulting architecture, illustrated in Figure~\ref{neuron_gating_activation} (d), represents the final searched model. In this example, one of the three neurons is pruned, while the remaining neurons are assigned their selected activation functions.
\end{enumerate}

Similar algorithms and corresponding visual representations, along with their interpretations, can be developed for NG- and MA-based NAS; however, these are omitted for brevity. Although the proposed formulations and algorithm are presented in the context of MLP architectures, the relaxed bilevel framework extends naturally to CNNs by associating architectural variables with convolutional channels and the final linear layers, and can further be adapted to search for optimal convolutional filters. Notably, the underlying optimization structure, relaxation strategy, and solution methodology remain unchanged. To avoid redundancy, detailed derivations for CNNs are omitted, as they follow directly from the MLP case with appropriate tensor generalizations.

\section{Experimental Results and Discussion}\label{sec5}
This section presents a comprehensive evaluation of the proposed relaxed bilevel formulations on MNIST and CIFAR-10 image classification tasks. We systematically analyze the performance of the proposed architecture search strategies---namely, NAS-NG, NAS-MA, and NAS-NGMA---on both MLP and CNN architectures, and compare them with their corresponding dense baseline models. To provide a broader context, we also compare our results with representative models from the literature on MNIST. Furthermore, the effectiveness of the proposed methods is validated by comparing them with DARTS on the CIFAR-10 dataset.

\subsection{Datasets and Baseline Models}
The MNIST dataset comprises 70,000 grayscale images of handwritten digits (0--9), each with a spatial resolution of $28 \times 28$ pixels, and is partitioned into 60,000 training samples and 10,000 test samples. Following standard preprocessing practices, pixel intensities are normalized to the $[0,1]$ range. For MLP-based models, each image is flattened into a 784-dimensional input vector. As a primary baseline, we consider a fully connected MLP corresponding to the maximum-capacity architecture induced by the proposed framework, wherein all neurons are active. This configuration serves as a reference model prior to any architectural adaptation through the proposed relaxation mechanisms. To ensure a comprehensive evaluation, multiple dense MLP architectures with varying depths and widths are examined. Specifically, we consider networks with 1 to 4 hidden layers, where each configuration employs a uniform number of neurons per layer. The hidden layer sizes are chosen from $\{1024, 2048, 3072\}$, resulting in a total of 12 baseline configurations. All models use ReLU activation functions in the hidden layers and a softmax output layer for the 10-class classification task. In addition to MLPs, we evaluate the proposed methods on convolutional architectures to assess their effectiveness in learning spatial features. Two CNN baseline architectures are considered. The first architecture comprises three convolutional layers with channel sizes [64, 128, 256] and kernel sizes [5, 3, 3], respectively. The second architecture consists of four convolutional layers with channel sizes [32, 64, 128, 256] and kernel sizes [5, 3, 3, 3]. Both architectures are followed by two fully connected layers with dimensions [256, 128]. 

Furthermore, we extend our evaluation to the CIFAR-10 dataset, which consists of 50,000 training images and 10,000 test images distributed across 10 classes. In contrast to MNIST, CIFAR-10 images are represented in RGB format with a spatial resolution of $32 \times 32$ pixels. For CIFAR-10, architecture search is performed using CNN models comprising three convolutional layers with channel sizes [128, 256, 512] and corresponding kernel sizes [5, 3, 3]. This feature extraction backbone is followed by a fully connected layer of dimension 512. To benchmark the effectiveness of the proposed method, we conduct architecture search experiments using the DARTS framework under similar experimental settings and compare the resulting architectures with those obtained by our approach. All models are trained using mini-batch gradient-based optimization, with detailed training hyperparameters provided in the subsequent section.

\subsection{Experimental Setup}
We evaluate the proposed methods in a constrained setting with a 2-day wall-time limit. All experiments are conducted on a single node of our High Performance Computing (HPC) cluster, equipped with sufficient memory for the configurations considered. The workflow consists of two phases: (i) architecture search using the relaxed bilevel formulation-based approaches, and (ii) retraining the discovered optimal architecture from scratch, followed by final evaluation on the test set. Both MLPs and CNNs are used for MNIST, whereas only CNNs are used for CIFAR-10.

A batch size of 64 is considered across all phases of experiments on the MNIST dataset. During the architecture search phase, we employ the Adam optimizer with a learning rate of $\eta_A = 3 \times 10^{-4}$ and a weight decay of $1 \times 10^{-3}$. The lower-level optimization problem is approximated using one-step gradient descent to compute the hypergradients. Unless otherwise specified, the architecture search is conducted for 10 epochs; however, for two computationally intensive experiments, the search is limited to 8 epochs. For architecture search, the available training data is evenly split into training and validation subsets. All architecture parameters are initialized from a standard normal distribution scaled by $10^{-3}$. After the search phase, the selected architecture is trained from scratch using mini-batch gradient descent. The training process uses an initial learning rate of 0.025 (with a minimum learning rate of $1 \times 10^{-3}$), momentum of 0.9, weight decay of $3 \times 10^{-4}$, and gradient clipping to ensure numerical stability. All final architectures are trained for 25 epochs before evaluation on the test set. 

We used fixed random seeds in the experiments. While conducting experiments on CNN architectures for MNIST, we used slightly different hyperparameter settings than those for MLPs. Specifically, the minimum learning rate for network training was set to $1 \times 10^{-4}$, and the architecture learning rate was set to $6 \times 10^{-4}$. For architecture evaluation, models were trained from scratch for 30 epochs in the case of CNNs with three convolutional layers and for 50 epochs in the case of CNNs with four convolutional layers. Additionally, a dropout rate of 0.15 was used during the architecture search phase, and 0.25 during the final evaluation phase. For CIFAR-10 experiments, the DARTS experimental setup closely follows the protocol established in the original work by \cite{DBLP:conf/iclr/LiuSY19}, with minor changes to the optimization hyperparameters, e.g., batch size of 96 and minimum learning rate of $2.5 \times 10^{-5}$. The DARTS architecture consists of eight cells: six normal cells and two reduction cells. The reduction cells are positioned at one-third and two-thirds of the network depth to progressively downsample the feature representations. Within the normal cells, all operations are performed with a stride of 1, thereby preserving the spatial resolution. In contrast, the reduction cells utilize a stride of 2 to halve the spatial dimensions of the feature maps. The set of candidate operations, denoted by $\mathcal{O}$, is identical to that defined in \cite{DBLP:conf/iclr/LiuSY19}. For proposed methods, we considered a three-convolutional-layer model as discussed earlier.

\subsection{MLP Model}
The results of architecture search and evaluation for MLP models on the MNIST dataset are presented as follows.
\subsubsection{Architecture Search}
The architecture search phase aims to identify optimal sub-networks from initially dense MLP configurations using three proposed search strategies: NAS-NG, NAS-MA, and their combination (NAS-NGMA). The NG approach follows a structured pruning strategy in which neuron importance is evaluated based on their contribution to the learned representation, and redundant neurons are progressively removed. The MA-based method analyzes neuron activation magnitudes across the training data and prunes those that consistently exhibit low responses over diverse inputs. The combined approach integrates both gating-based importance measures and activation statistics to enable more stable pruning decisions, balancing model compactness and predictive performance.

The architecture search process iteratively refines networks based on validation performance until either the preset number of iterations is reached or the allocated computational budget of two days is exhausted. The final architecture is selected based on the minimum validation loss, while validation accuracy is monitored as a secondary metric. Table~\ref{tab:mlp_search} summarizes the validation performance during architecture search, along with the corresponding execution time and resulting model size, starting from the baseline model, which has a model size of $8.97 \pm 9.29$ million parameters. Notable differences are observed in both search time and model size across the considered architecture search methods.
\begin{table}[ht]
\centering
\caption{Architecture search performance for MLP models on the MNIST dataset.}
\label{tab:mlp_search}
\resizebox{0.65\linewidth}{!}{
\begin{tabular}{lcccccc}
\hline
\multicolumn{7}{c}{\textbf{1--4 Hidden Layers $\times$ \{1024, 2048, 3072\} Neurons/Layer}} \\
\textbf{Search Method} 
& \multicolumn{2}{c}{\textbf{Val Acc (\%)}} 
& \multicolumn{2}{c}{\textbf{Time (hrs)}} 
& \multicolumn{2}{c}{\textbf{Params (M)}} \\
& \textbf{Mean} & \textbf{Std Dev} 
& \textbf{Mean} & \textbf{Std Dev} 
& \textbf{Mean} & \textbf{Std Dev} \\
\hline
NAS-NG  & 97.59 & 0.04 & 0.45 & 0.14 & 1.90 & 1.33 \\
NAS-MA  & 97.27 & 0.12 & 18.80 & 12.04 & 4.64 & 4.38 \\
NAS-NGMA & 96.20 & 0.31 & 18.81 & 11.89 & 3.54 & 3.56 \\
\hline
\end{tabular}
}
\end{table}

\begin{figure}[ht]
\centering
\includegraphics[width=0.9\textwidth]{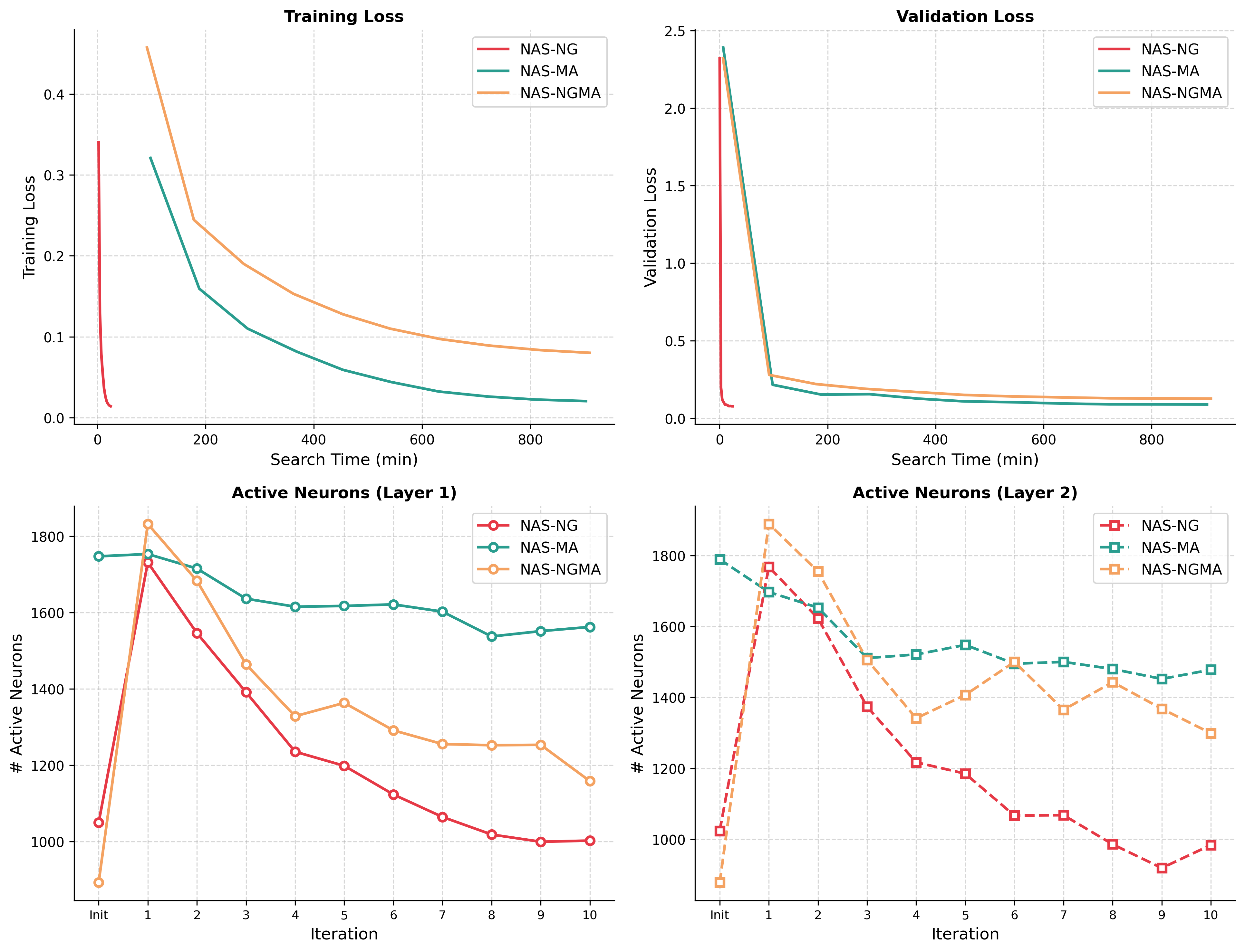}
\caption{Comparison of architecture search methods for MLP models on the MNIST dataset.}
\label{fig:arch_search_mlp}
\end{figure}

The NAS-NG method demonstrates remarkable efficiency, completing the search in an average of 0.45 CPU hours—approximately 40 times faster than the MA-based approaches (NAS-MA and NAS-NGMA). This substantial speed advantage stems from the computational simplicity of gating-based importance metrics, which can be evaluated more rapidly than comprehensive activation pattern analysis. The MA-based methods require extensive forward passes through the network with diverse input data to capture representative activation statistics, resulting in mean search times of approximately 18.8 hours. Further, the standard deviation in search times reflects the varying complexity of different initial configurations. Deeper networks with more neurons naturally require longer search durations, as evidenced by the maximum search times approaching 40 hours for the 4-layer, 3072-dimensional configurations under MA-based search. Figure~\ref{fig:arch_search_mlp} depicts the search performance of the proposed NAS methods for MLP architectures on the MNIST dataset, along with the layer-wise evolution of active neurons during the course of optimization.

\subsubsection{Architecture Evaluation}
Following the search phase, all discovered architectures are retrained from scratch to evaluate their generalization performance on the test set. This evaluation phase constitutes the most critical comparison, as it reflects the practical effectiveness of the searched architectures when trained independently of the search procedure. Table~\ref{tab:overall_performance_mlp} summarizes the model complexity and test accuracy metrics across all methods for the MLP architectures. The findings indicate that the models obtained using the suggested architecture search methods achieve performance very close to the baseline while also attaining substantial parameter reductions.
\begin{table}[ht]
\centering
\caption{Test performance of discovered MLP models on MNIST dataset.}
\label{tab:overall_performance_mlp}
\resizebox{0.75\linewidth}{!}{
\begin{tabular}{lcccc}
\hline
\textbf{Search Method} & \textbf{Mean Test Acc.} & \textbf{Max Test Acc.} & \textbf{Mean Params} & \textbf{Param} \\
                      & \textbf{(\%)}      & \textbf{(\%)}     & \textbf{(M)}        & \textbf{Reduction (\%)} \\
\hline
Baseline      & 98.56 & 98.69 & 8.97 & --- \\
NAS-NG            & 98.53 & 98.60 & 1.90 & 78.8 \\
NAS-MA            & 98.50 & 98.65 & 4.64 & 48.3 \\
NAS-NGMA          & 98.55 & 98.68 & 3.54 & 60.5 \\
\hline
\end{tabular}
}
\end{table}

A detailed configuration-wise performance analysis for the MLP architectures is provided in Appendix~\ref{config_specific}, where heatmap visualizations illustrate the performance of the proposed search methods across different network depths and hidden dimensions.

\subsection{CNN Model}
The experiments for CNN models are divided into two categories based on the datasets used, namely: experiments on MNIST and experiments on CIFAR-10, as follows.
\subsubsection{MNIST}
The results of architecture search and evaluation for CNN models on the MNIST dataset are presented as follows.\\

\begin{figure}[ht]
\centering
\includegraphics[width=0.9\textwidth]{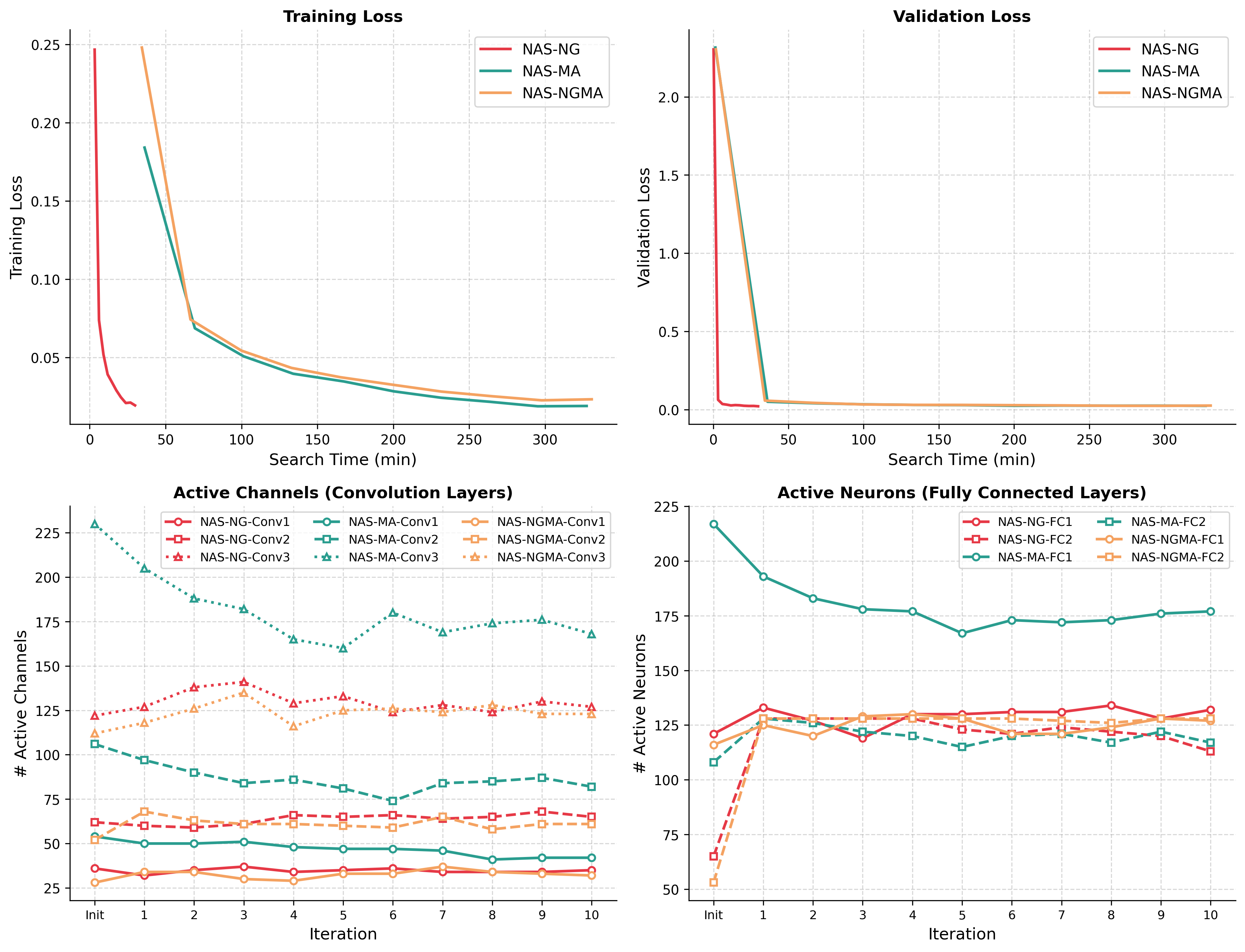}
\caption{Comparison of architecture search methods for CNN models on the MNIST dataset.}
\label{fig:arch_search_cnn_mnist}
\end{figure}

\noindent \textbf{Architecture Search.}\\
Figure~\ref{fig:arch_search_cnn_mnist} depicts the search performance of the proposed NAS methods for CNN architectures on the MNIST dataset, along with the layer-wise evolution of active channels in the convolutional layers and active neurons in the fully connected layers over the course of optimization.

\begin{table}[ht]
\centering
\caption{Architecture search performance for CNN models on the MNIST dataset.}
\label{tab:cnn_search_mnist}
\resizebox{0.85\linewidth}{!}{
\begin{tabular}{lcccccc}
\hline
\textbf{Search Method} 
& \multicolumn{3}{c}{\textbf{3 Layer Conv [64, 128, 256]}} 
& \multicolumn{3}{c}{\textbf{4 Layer Conv [32, 64, 128, 256]}} \\
& \textbf{Val Acc (\%)} & \textbf{Time (hrs)} & \textbf{Params (M)} 
& \textbf{Val Acc (\%)} & \textbf{Time (hrs)} & \textbf{Params (M)} \\
\hline
NAS-NG  & 99.31 & 0.50 & 0.26 & 99.24 & 0.44 & 0.12 \\
NAS-MA & 99.22 & 5.46 & 0.45 & 99.24 & 4.64 & 0.24 \\
NAS-NGMA  & 99.26 & 5.51 & 0.25 & 99.24 & 4.59 & 0.14 \\
\hline
\end{tabular}
}
\end{table}

A summary of the architecture search results for CNN models on the MNIST dataset, starting from the baseline models with model sizes of $1.00$ and $0.49$ million parameters, is presented in Table~\ref{tab:cnn_search_mnist}. Notably, CNN-based architecture search requires significantly less time than MLP-based search in the case of activation-driven NAS methods. This table also suggests that the NAS-NG method’s rapid pruning decisions, though based on simpler criteria, are remarkably effective at identifying promising architectures, consistent with observations in the case of MLPs.\\

\noindent \textbf{Architecture Evaluation.}\\
Table~\ref{tab:overall_performance_cnn} presents the model complexity and test accuracy metrics for the two convolutional layer configurations considered. The findings indicate that the models obtained using the suggested architecture search methods achieve performance that is either higher than or very close to the baseline. They also attain substantial parameter reductions, similar to those observed in the case of MLP models.
\begin{table}[ht]
\centering
\caption{Test performance of discovered CNN models on MNIST dataset.}
\label{tab:overall_performance_cnn}
\resizebox{0.715\linewidth}{!}{
\begin{tabular}{lcccc}
\hline
\multirow{2}{*}{\textbf{Search Method}} 
& \multicolumn{2}{c}{\textbf{3 Layer Conv [64, 128, 256]}} 
& \multicolumn{2}{c}{\textbf{4 Layer Conv [32, 64, 128, 256]}} \\
& \textbf{Test Acc (\%)} 
& \textbf{Params $\downarrow$ (\%)} 
& \textbf{Test Acc (\%)} 
& \textbf{Params $\downarrow$ (\%)} \\
\hline
Baseline  & 99.60 & -    & 99.62 & -    \\
NAS-NG        & 99.63 & 73.5 & 99.50 & 74.7 \\
NAS-MA        & 99.49 & 55.1 & 99.55 & 50.5 \\
NAS-NGMA       & 99.61 & 75.4 & 99.56 & 72.3 \\
\hline
\end{tabular}
}
\end{table}

\subsubsection{CIFAR-10}
The results of architecture search and evaluation for CNN models on the CIFAR-10 dataset are presented as follows. In addition to the proposed NAS methods, architecture search experiments are also conducted using the DARTS method for comparison purposes.\\

\begin{figure}[ht]
\centering
\includegraphics[width=0.9\textwidth]{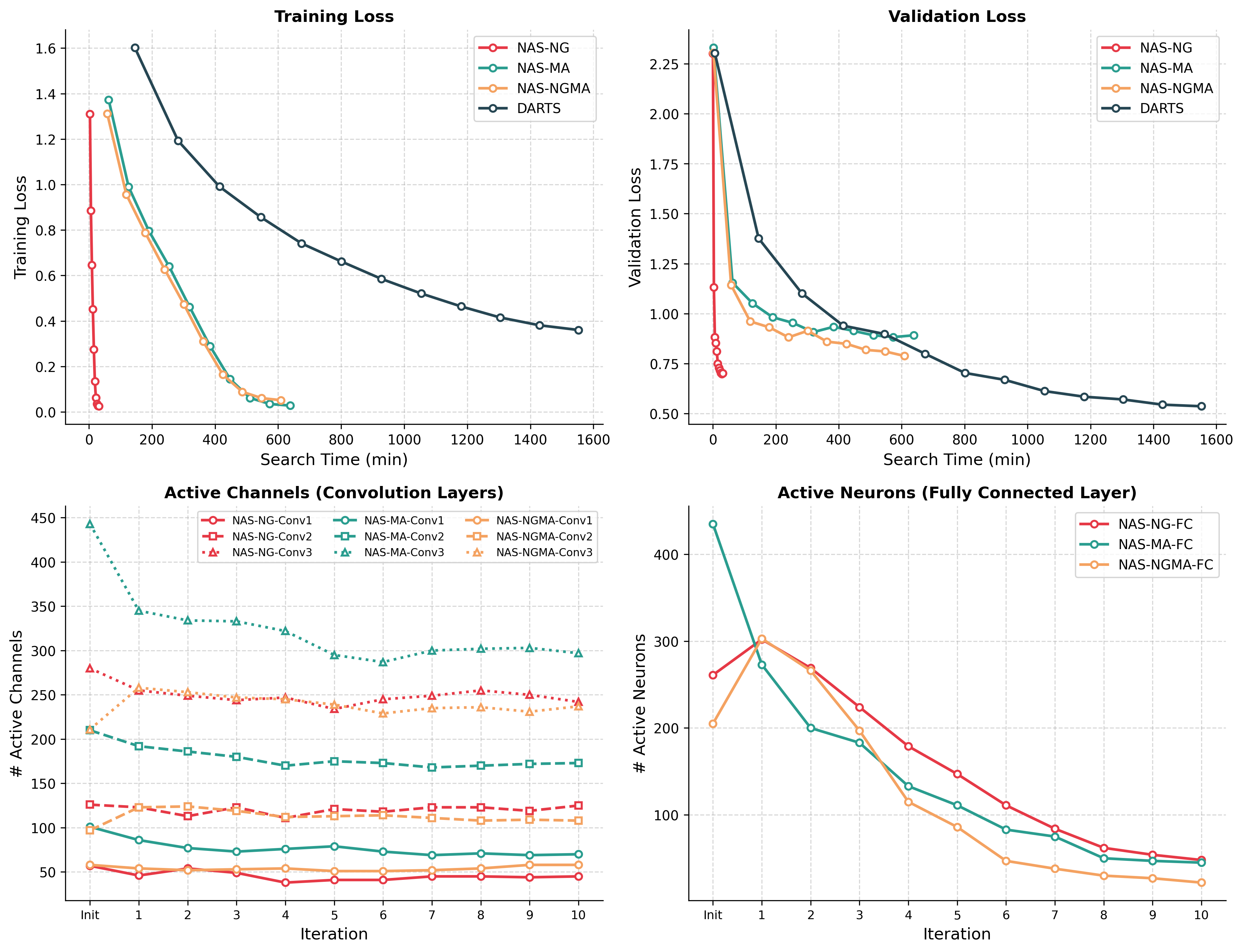}
\caption{Comparison of NAS methods for CNN models on the CIFAR-10 dataset.}
\label{fig:arch_search_cnn_cifar10}
\end{figure}

\noindent \textbf{Architecture Search.}\\
Table~\ref{tab:cnn_search_cifar10} presents a summary of the architecture search results for CNN models on the CIFAR-10 dataset. The proposed methods start from a baseline model with $5.69$ million parameters. Figure~\ref{fig:arch_search_cnn_cifar10} illustrates the search performance of the proposed NAS methods and the DARTS method for CNN architectures on the CIFAR-10 dataset, along with the layer-wise evolution of active channels in the convolutional layers and active neurons in the fully connected layer.\\

\begin{table}[ht]
\centering
\caption{Architecture search performance for CNN models on the CIFAR-10 dataset.}
\label{tab:cnn_search_cifar10}
\resizebox{0.65\linewidth}{!}{
\begin{tabular}{lcccccc}
\hline
\textbf{Search Method} 
& \multicolumn{2}{c}{\textbf{Val Acc (\%)}} 
& \multicolumn{2}{c}{\textbf{Time (hrs)}} 
& \multicolumn{2}{c}{\textbf{Params (M)}} \\
& \textbf{Mean} & \textbf{Std Dev} 
& \textbf{Mean} & \textbf{Std Dev} 
& \textbf{Mean} & \textbf{Std Dev} \\
\hline
NAS-NG  & 77.59 & 0.31 & 0.56 & 0.02 & 0.49 & 0.04 \\
NAS-MA  & 74.16 & 0.13 & 10.75 & 0.39 & 0.83 & 0.10 \\
NAS-NGMA & 74.88 & 0.24 & 10.19 & 0.08 & 0.41 & 0.03 \\
DARTS & 81.93 & 0.31 & 26.30 & 0.56 & 3.40 & 0.10 \\
\hline
\end{tabular}
}
\end{table}

\noindent \textbf{Architecture Evaluation.}\\
Table~\ref{tab:overall_performance_cnn_cifar10} presents the test accuracy metrics and the reduction in model complexity for a three-layer convolutional configuration with one fully connected layer. The findings indicate that the models obtained using the proposed methods perform better than the architectures discovered by the vanilla DARTS method while discovering architectures with substantially smaller network sizes.

\begin{table}[ht]
\centering
\caption{Test performance of discovered CNN models on CIFAR-10 dataset.}
\label{tab:overall_performance_cnn_cifar10}
\resizebox{0.6\linewidth}{!}{
\begin{tabular}{lccccc}
\hline
\textbf{Search Method} 
& \multicolumn{2}{c}{\textbf{Test Acc (\%)}} 
& \multicolumn{2}{c}{\textbf{Eval. Time (hrs)}} 
& \multicolumn{1}{c}{\textbf{Params $\downarrow$ (\%)}} \\
& \textbf{Mean} & \textbf{Std Dev} 
& \textbf{Mean} & \textbf{Std Dev} 
&  \\
\hline
Baseline  & 86.14 & 0.22 & 1.48 & 0.05 & --  \\
NAS-NG        & 85.60 & 0.21 & 1.09 & 0.02 &  91.4  \\
NAS-MA        & 84.55 & 0.31 & 4.61 & 0.22 &  85.4  \\
NAS-NGMA       & 84.73 & 0.18 & 3.08 & 0.08 &  92.8 \\
DARTS     & 82.03 & 0.50 & 20.20 & 0.47 & -- \\
\hline
\end{tabular}
}
\end{table}

\subsection{Overall Performance Comparison}
The above findings indicate that the proposed architecture search methods can either match or come very close to the baseline performance while achieving substantial parameter reductions across both types of architectures on the MNIST and CIFAR-10 datasets. On CIFAR-10, their performance outperforms the vanilla DARTS method.

Several notable observations emerge from this comparison for MLPs. First, all three search methods achieve mean test accuracies within 0.06 percentage points of the baseline, indicating that the discovered architectures retain the essential representational capacity needed for MNIST classification. The NAS-NG approach achieves the most impressive parameter reduction, reducing the average model size from 8.97M to 1.90M while maintaining 98.53\% accuracy. The NG, when combined with a MA-based method, achieves the highest maximum accuracy (98.68\%), matching the best baseline performance while using 60.5\% fewer parameters. This suggests that careful coordination between multiple pruning criteria can occasionally identify architectures that achieve the full performance of dense networks while operating at a fraction of the computational cost. The MA-based NAS method falls in the middle ground, achieving a 48.3\% reduction in parameters. While still substantial, this more conservative pruning reflects the method's tendency to retain neurons with moderate activation patterns that may contribute to robustness, even if not strictly necessary for high validation accuracy. Similar performance improvement and complexity reduction trends are observed for CNNs.

\subsubsection{Accuracy vs Model Complexity}
\begin{figure}[ht]
\centering
\includegraphics[width=\textwidth]{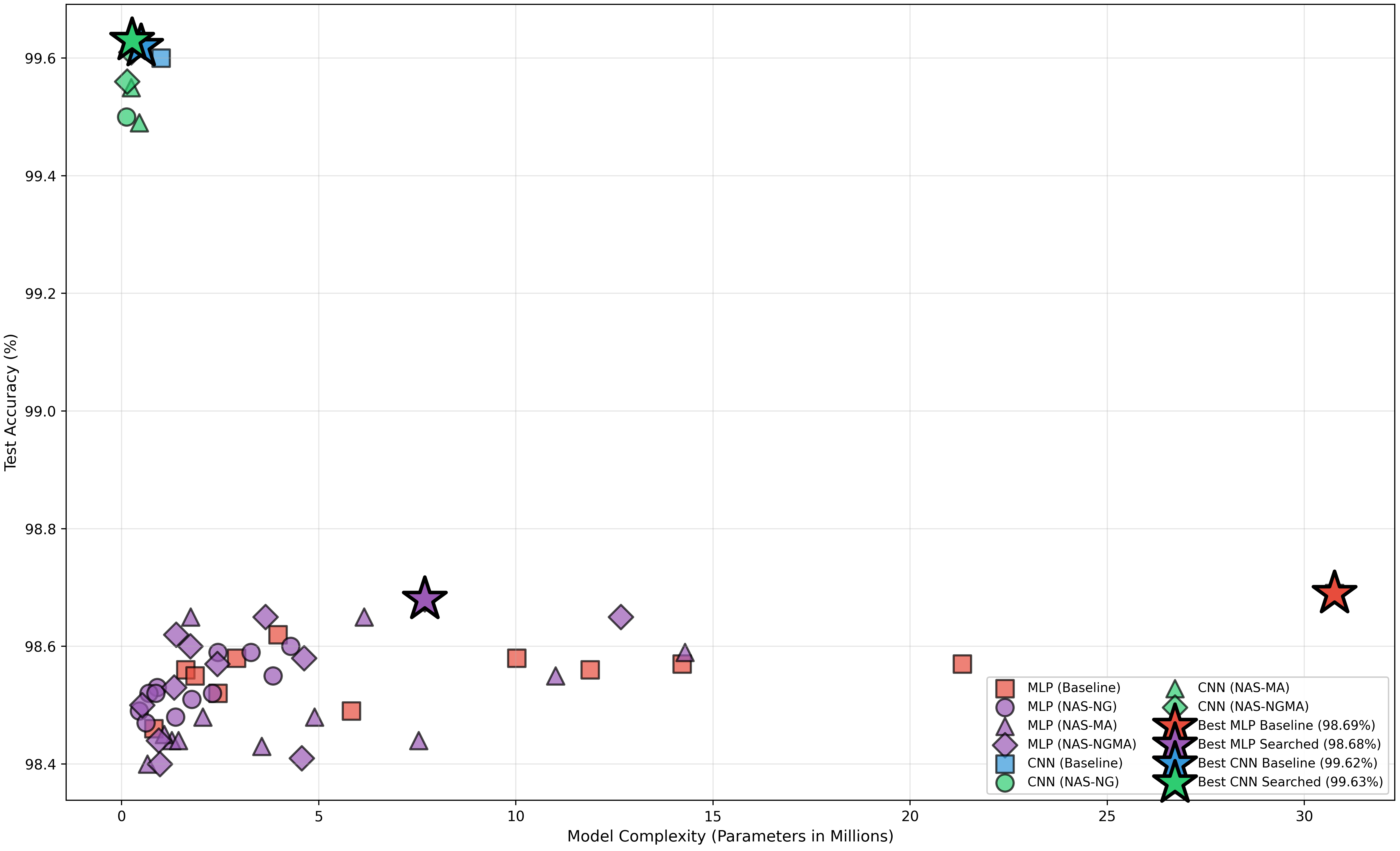}
\caption{Test accuracy vs model complexity (measured in millions of parameters) for all evaluated configurations on the MNIST dataset.}
\label{fig:acc_complexity}
\end{figure}

Figure~\ref{fig:acc_complexity} presents a scatter plot revealing the fundamental trade-off between model complexity and test accuracy.
This visualization clearly demonstrates the advantage of architecture search methods. Baseline models have higher parameter counts, while search methods achieve comparable accuracy with substantially fewer parameters. The best baseline MLP model (98.69\% accuracy, 30.76M parameters) and best MLP searched model (98.68\% accuracy, 7.69M parameters) are highlighted. The baseline MLP configurations span a wide range of parameter counts from 0.81M to 30.76M, with test accuracies between 98.46\% and 98.69\%. In contrast, the searched MLP architectures occupy the lower-left region of the plot, achieving similar accuracies with dramatically reduced parameter counts. This clustering pattern indicates that architecture search successfully identifies the minimal architectural complexity needed to solve the task effectively.

The best-performing architecture, discovered using the combined gating and mixed activation method, achieves 98.68\% accuracy with only 7.69M parameters—a 75\% reduction compared to the best baseline configuration (30.76M parameters at 98.69\% accuracy). This near-identical performance with a quarter of the parameters represents a substantial improvement in model efficiency. The baseline CNN models generally exhibit smaller parameter sizes than their MLP counterparts while achieving strong test accuracy. However, the CNN architectures obtained through the proposed search methods demonstrate further improvements, yielding significantly reduced model complexity and, in some cases, outperforming even the best-performing baseline models, as shown in Table~\ref{tab:overall_performance_cnn}.

\subsubsection{Additional Analysis}
A detailed analysis of the effects of network depth, hidden dimensions, and parameter reduction achieved by the proposed search methods is provided in Appendix~\ref{add_analy}. In particular, the appendix examines the relationship between architectural capacity and classification performance across different MLP configurations, along with the compression characteristics of the proposed NAS strategies. The results indicate that the search methods maintain competitive accuracy across varying depths and hidden dimensions while significantly reducing model complexity. Additional quantitative comparisons and visualizations are also provided.

\subsection{Comparative Analysis}
This section compares the models discovered by the NAS methods. We also include selected results from the literature on the MNIST dataset. Notably, no clear positive correlation is observed between search duration and final test accuracy. For instance, the NAS-NG method achieves search times of under one hour while attaining accuracies spanning the full observed range (98.47\% to 98.60\%) on the MNIST dataset. In contrast, the MA-based methods, despite requiring 20--40 hours of search, achieve accuracies that are comparable to, and in some cases lower than, those obtained by simpler methods. This observation has important practical implications. It suggests that computationally intensive and sophisticated search algorithms are not always necessary to discover effective architectures, particularly for datasets of moderate complexity such as MNIST. 

\subsubsection{Best Performing Architectures}
Table~\ref{tab:best_models} presents the optimal MLP configuration discovered by each method, along with the best baseline for reference.
\begin{table}[ht]
\centering
\caption{MLP best performing models identified by each search method on MNIST dataset.}
\label{tab:best_models}
\resizebox{0.7\linewidth}{!}{
\begin{tabular}{lcccccc}
\hline
\textbf{Method} & \textbf{Layers} & \textbf{Config.} & \textbf{Test Acc.} & \textbf{Params} & \textbf{Search} \\
               &                &                 & \textbf{(\%)}      & \textbf{(M)}    & \textbf{Time (hrs)} \\
\hline
Baseline   & 4 & [3072, 3072, 3072, 3072] & 98.69 & 30.76 & --- \\
NAS-NG         & 4 & [3072$^4$] $\rightarrow$ Pruned & 98.60 & 4.29  & 0.76 \\
NAS-MA         & 4 & [1024$^4$] $\rightarrow$ Pruned & 98.65 & 1.75  & 16.16 \\
NAS-NGMA        & 3 & [3072$^3$] $\rightarrow$ Pruned & 98.68 & 7.69  & 35.80 \\
\hline
\end{tabular}
}
\end{table}
The NGMA-based NAS method achieves the highest accuracy among search methods at 98.68\%, falling just 0.01 percentage points short of the best baseline while using 75\% fewer parameters. Interestingly, this top configuration emerges from a 3-layer rather than a 4-layer network, supporting the earlier observation that interactions among pruning criteria may favor intermediate depths. The NAS-MA method produces the most compact high-performing model, achieving 98.65\% with only 1.75M parameters—a remarkable 94.3\% reduction compared to the best baseline. This configuration began with the smallest dense architecture (4 layers × 1024 neurons) and optimized it via activation analysis, demonstrating that starting with smaller initial networks can sometimes yield excellent results when combined with effective search. The NAS-NG method's best configuration achieves 98.60\% with 4.29M parameters (an 86\% reduction), demonstrating its ability to dramatically compress large networks while retaining strong performance.
\begin{table}[ht]
\centering
\caption{CNN best performing models identified by each search method on MNIST dataset.}
\label{tab:cnn_best_models}
\resizebox{0.72\linewidth}{!}{
\begin{tabular}{lcccccc}
\hline
\textbf{Method} & \textbf{Layers} & \textbf{Config.} & \textbf{Test Acc.} & \textbf{Params} & \textbf{Search} \\
               &                &                 & \textbf{(\%)}      & \textbf{(M)}    & \textbf{Time (hrs)} \\
\hline
Baseline       & 4 & [32, 64, 128, 256] & 99.62 & 0.49 & --- \\
NAS-NG             & 3 & [64, 128, 256] $\rightarrow$ Pruned & 99.63 & 0.26  & 0.50 \\
NAS-MA             & 4 & [32, 64, 128, 256] $\rightarrow$ Pruned & 99.55 & 0.24  & 4.64 \\
NAS-NGMA            & 3 & [64, 128, 256] $\rightarrow$ Pruned & 99.61 & 0.25  & 5.51 \\
\hline
\end{tabular}
}
\end{table}

In the case of CNNs on the MNIST dataset, the best-performing model is obtained using the NAS-NG method with a base architecture consisting of three convolutional layers. This model outperforms the best CNN baseline by 0.01\% while being approximately 47\% less complex in terms of model size. The best-performing CNN models on the MNIST dataset are presented in Table~\ref{tab:cnn_best_models}. For the CIFAR-10 dataset, the best obtained models are provided in Table~\ref{tab:cnn_best_models_cifar10}.

\begin{table}[ht]
\centering
\caption{CNN best performing models identified by each search method on CIFAR-10 dataset.}
\label{tab:cnn_best_models_cifar10}
\resizebox{0.85\linewidth}{!}{
\begin{tabular}{lcccccc}
\hline
\textbf{Method} & \textbf{Layers/Cells} & \textbf{Config.} & \textbf{Test Acc.} & \textbf{Params} & \textbf{Search} \\
               &                &                 & \textbf{(\%)}      & \textbf{(M)}    & \textbf{Time (hrs)} \\
\hline
Baseline       & 3 & [128, 256, 512] & 86.34 & 5.69 & --- \\
NAS-NG             & 3 & [128, 256, 512] $\rightarrow$ Pruned & 85.77 & 0.51  & 0.53 \\
NAS-MA             & 3 & [128, 256, 512] $\rightarrow$ Pruned & 84.83 & 0.79 & 10.64 \\
NAS-NGMA            & 3 & [128, 256, 512] $\rightarrow$ Pruned & 84.85 & 0.38  & 10.15 \\
DARTS          & 20 & Stacked Normal--Reduction Cells  & 82.57 & 3.38 & 25.88 \\
\hline
\end{tabular}
}
\end{table}

\subsubsection{Efficiency Metrics}
\begin{figure}[ht]
\centering
\includegraphics[width=0.9\textwidth]{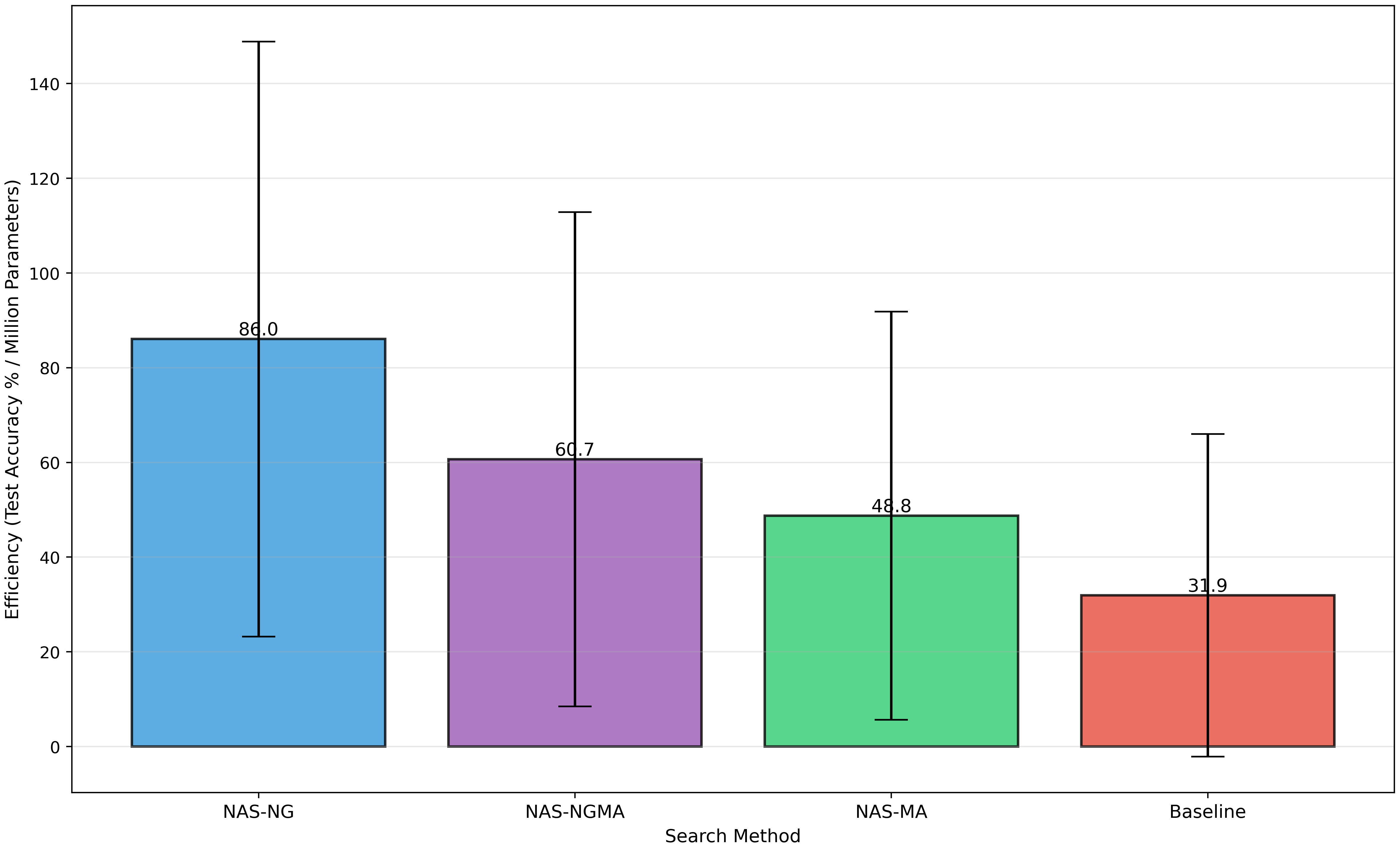}
\caption{Mean efficiency (test accuracy per million parameters) achieved by each search method for MLP models on the MNIST dataset.}
\label{fig:efficiency}
\end{figure}

To provide a unified measure of architecture quality, we computed an efficiency metric defined as test accuracy percentage divided by parameter count in millions. Figure~\ref{fig:efficiency} presents this comparison. Higher values indicate better accuracy-to-parameter ratios. Error bars represent standard deviation across all configurations. NAS-NG achieves the highest efficiency, delivering 86.05 accuracy points per million parameters.

The NAS-NG method achieves the highest mean efficiency at 86.05 accuracy points per million parameters—2.7 times higher than the baseline (31.94) and 1.8 times higher than the MA-based NAS method (48.75). This metric quantifies the intuition visible in earlier plots: NAS-NG discovers architectures that achieve near-baseline accuracy with dramatically reduced complexity.

The NAS-NGMA method achieves an efficiency of 60.66, representing a middle ground between the aggressive compression of NAS-NG and the more conservative approach of MA. The substantial standard deviations (particularly for NAS-NG at 62.80) reflect the diversity of configurations tested; some initial networks offer more redundancy to exploit than others. The five most efficient individual models on MNIST and CIFAR-10 datasets are presented in Table~\ref{tab:top_efficient}. Remarkably, the most efficient model is discovered by the NAS-NG method. These models require significantly less memory and computational resources.
\begin{table}[ht]
\centering
\caption{Top five most efficient models (highest accuracy-to-parameter ratios) on MNIST and CIFAR-10 datasets.}
\label{tab:top_efficient}
\resizebox{0.73\linewidth}{!}{
\begin{tabular}{lccc ccc}
\hline
\textbf{Rank} 
& \multicolumn{3}{c}{\textbf{MNIST}} 
& \multicolumn{3}{c}{\textbf{CIFAR-10}} \\
& \textbf{Method} & \textbf{Config.} & \textbf{Efficiency} 
& \textbf{Method} & \textbf{Config.} & \textbf{Efficiency} \\
\hline
1 & NAS-NG & [32, 64, 128, 256] & 803.97 & NAS-NGMA & [128, 256, 512] & 225.65 \\
2 & NAS-NG & [32, 64, 128, 256] & 803.89 & NAS-NGMA & [128, 256, 512] & 199.23 \\
3 & NAS-NGMA & [32, 64, 128, 256] & 732.76 & NAS-NG & [128, 256, 512] & 195.28 \\
4 & NAS-NGMA & [32, 64, 128, 256] & 732.32 & NAS-NGMA & [128, 256, 512] & 193.98 \\
5 & NAS-MA & [32, 64, 128, 256] & 410.58 & NAS-NG & [128, 256, 512] & 166.88 \\
\hline
\end{tabular}
}
\end{table}

\subsubsection{Comparative Evaluation on the MNIST Benchmark}
We present a comparative evaluation of the proposed architecture search strategies against established models reported in the literature on the MNIST benchmark dataset. The results in Table~\ref{tab:mnist_comparison} demonstrate that the searched architectures achieve competitive performance relative to several widely adopted deep learning and classical approaches, while maintaining simpler model structures, as discussed earlier.
\begin{table}[ht]
\centering
\caption{Comparison of MNIST classification accuracies from various models.}
\label{tab:mnist_comparison}
\resizebox{0.93\linewidth}{!}{
\begin{tabular}{l c l c}
\hline
\textbf{Model} & \textbf{Test Acc. (\%)} & \textbf{Model} & \textbf{Test Acc. (\%)} \\
\hline
CNN \cite{kaziha2019comparison} & 99.45  & LSTM \cite{kaziha2019comparison} & 99.22\\
LeNet-5 \cite{lecun2002gradient} & 99.05 & Consensus Clustering \cite{rexy2019handwritten} & 95 \\
LR-based \cite{kang2018new} & 93.83 & SVM \cite{reddy2022handwritten} & 97.83 \\
Spiking-CNN \cite{kalbande2022performance} & 96.89 & Multi-Column DNN \cite{ciregan2012multi} & 99.77  \\
PCA/LDA \cite{patel2019handwritten} & 86.60 & EfficientDet-D4 \cite{ahmed2023novel} & 99.83 \\
CNN-SVM \cite{ullah2025handwritten} & 99.30 &  RMDL \cite{kowsari2018rmdl} & 99.82\\
DropConnect Regularization \cite{wan2013regularization} & 99.79 &  APAC \cite{sato2015apac} & 99.77\\
Baseline (MLP) & 98.69 & Baseline (CNN) & 99.62 \\
NAS-NG (MLP) & 98.60 &  NAS-NG (CNN) & 99.63\\
NAS-MA (MLP) & 98.65 &  NAS-MA (CNN) & 99.55\\
NAS-NGMA (MLP) & 98.68 &  NAS-NGMA (CNN) & 99.61\\

\hline
\end{tabular}
}
\end{table}

\subsection{Additional Experiments for NAS-NG}
To further obtain better models, we perform two additional experiments. First, we investigate whether NAS-NG remains effective when initialized from substantially over-parameterized architectures. Second, we examine whether the proposed NG strategy can further improve architectures that have already been optimized using state-of-the-art NAS methods.

\subsubsection{Larger Baselines}
To assess the scalability of NAS-NG, we construct significantly larger MLP and CNN baseline architectures. NAS-NG is then applied to each baseline to identify compact architectures. As shown in Table~\ref{tab:large_baseline_nasng}, NAS-NG consistently reduces the number of trainable parameters while maintaining, and in several cases improving, the final test accuracy. The results obtained from these larger baselines are superior to those reported earlier, suggesting that starting from larger architectures can lead to better final models. This is likely because NAS-NG has greater flexibility to prune redundant neurons without excessively compressing the network, thereby discovering architectures that are both more accurate and substantially smaller than their corresponding baselines.

\begin{table}[ht]
\centering
\caption{Evaluation experiments for NAS-NG with larger baseline architectures.}
\label{tab:large_baseline_nasng}
\resizebox{0.9\textwidth}{!}{
\begin{tabular}{llcccc}
\hline
\textbf{Model} & \textbf{Architecture} &
\multicolumn{2}{c}{\textbf{Baseline}} &
\multicolumn{2}{c}{\textbf{NAS-NG}} \\
\cline{3-6}
& & \textbf{Test Acc. (\%)} & \textbf{Params (M)}
& \textbf{Test Acc. (\%)} & \textbf{Params (M)} \\
\hline

\multicolumn{6}{c}{\textbf{MLP}} \\
\hline
& [3072, 3072, 3072, 3072, 3072, 3072]
& 98.68 & 49.64
& 98.65 & 4.80 \\

& [4096, 4096, 4096, 4096, 4096, 4096]
& 98.64 & 87.16
& 98.67 & 7.38 \\

& [6144, 6144, 6144, 6144, 6144, 6144]
& 98.64 & 193.66
& \textbf{98.73} & 12.97 \\

\hline
\multicolumn{6}{c}{\textbf{CNN Backbone: [128, 256, 512, 512, 1024]}} \\
\hline
& MLP Head: [4096, 2048]
& 88.45 & 84.11
& 89.38 & 19.37 \\

& MLP Head: [2048, 1024]
& 88.66 & 44.24
& \textbf{89.67} & 10.39 \\

& MLP Head: [1024, 512]
& 88.86 & 25.88
& 89.22 & 6.13 \\

\hline
\end{tabular}
}
\end{table}

\subsubsection{Better Initialization}
\begin{figure}[ht]
\centering
\includegraphics[width=0.675\linewidth]{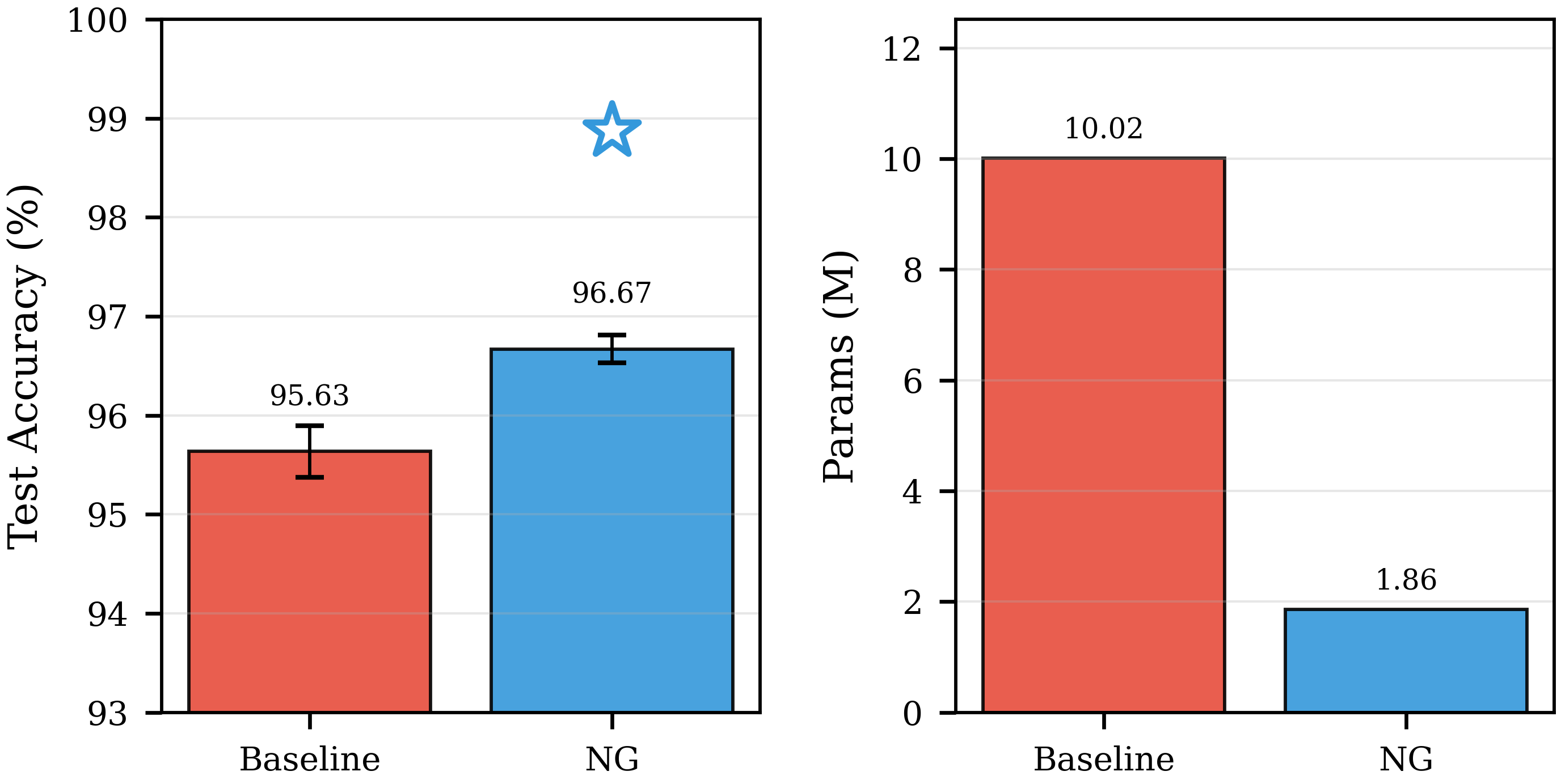}
\caption{Applying NG to literature-optimal DARTS cells simultaneously improves classification accuracy and reduces the number of trainable parameters, demonstrating that neuron-level optimization complements topology search.}
\label{fig:darts_breakdown}
\end{figure}

Previous experiments compared NAS-NG against architectures obtained from different initializations. Here, we investigate whether NG can provide further improvements when starting from architectures that have already been optimized by established NAS methods. Specifically, we apply NAS-NG to literature-optimal DARTS architectures while allowing only neuron-level optimization. During initialization, the optimal cell structures are kept unchanged, and only the number of stacked cells is varied to construct complete network architectures (10-cell and 14-cell macro-architectures). As shown in Figure~\ref{fig:darts_breakdown}, NAS-NG simultaneously improves classification accuracy and substantially reduces the number of trainable parameters. These results demonstrate that neuron-level optimization is complementary to topology search and can further refine architectures discovered by existing NAS methods.

\subsection{Discussion, Implications, and Recommendations}
The experimental findings highlight the effectiveness of the proposed NAS techniques in designing MLP and CNN architectures for MNIST classification, as well as CNN architectures for the CIFAR-10 classification task. All three approaches---NAS-NG, NAS-MA, and NAS-NGMA---identified architectures that maintained competitive accuracy (98.40\%--98.68\%) while achieving parameter reductions of 40\%--69\% for MLPs. For CNNs, the searched architectures achieved test accuracies of 99.49\%--99.63\% while attaining parameter reductions of 50.5\%--75.4\% on MNIST dataset. These findings extend NAS literature on network redundancy to fully connected architectures, showing that even simple search heuristics can remove excess capacity without performance loss.

For MLPs, among the methods, NAS-NG provides the best practical trade-off, delivering up to 86\% compression in under one hour with minimal hyperparameter tuning. This makes it highly appropriate for environments with limited resources, such as edge deployment. The MA-based method is more conservative (retaining $\sim$60\% parameters), potentially improving robustness to distribution shifts through moderate activation patterns; however, its 40$\times$ longer search time reduces cost-effectiveness. The combined approach achieves peak performance, matching baseline accuracy with 75\% fewer parameters through multi-criteria pruning. Nevertheless, its inconsistent average gains and longer search times limit broad applicability. Notably, no strong correlation is observed between network size (0.45M--30.76M parameters) and accuracy, suggesting that MNIST's low intrinsic complexity allows many architectures to achieve near-optimal performance once a minimum capacity threshold is exceeded.

Overall, these results question the necessity of computationally intensive search procedures. The sub-hour runtime of NAS-NG enables rapid iteration without extensive HPC resources, thereby democratizing NAS. Although MNIST’s simplicity limits overgeneralization, the findings underscore the importance of architecture search in identifying efficient operating points under diverse deployment constraints. The main implications and practical recommendations are summarized below.

\subsubsection{Implications for Neural Network Design}

These findings contribute to the discussion on designing neural networks. The reduction of parameters by 86\% (MLP) and 75.4\% (CNN) without losing accuracy shows that many networks have too many parameters. This aligns with the notion that more straightforward networks can perform equally well, as demonstrated in other research on network pruning and the lottery ticket hypothesis \cite{frankle2018lottery}.

The fact that simple search methods can perform as well as complex ones suggests that finding the best network design does not always require complicated algorithms. Although methods such as NAS via RL or DARTS are popular, our results show that simpler, problem-specific methods can be equally effective and require less computing power.

Lastly, the small accuracy ranges (98.40\% to 98.69\%) and (99.49\% to 99.63\%) across different setups suggests that for many uses, factors such as efficiency and ease of use might be more important than small differences in accuracy. This means examining the overall quality of a model, not just its accuracy.

\subsubsection{Practical Recommendations}
Based on our experiments, we recommend the following
\begin{enumerate}[label=(\alph*)]
\item \textbf{Start with neuron gating.} For most tasks, especially those with limited computing power or tight deadlines, NAS-NG is a good first choice. It is quick and effectively reduces the parameters.

\item \textbf{Consider task complexity.} For tasks more complex than MNIST, methods that prune more carefully might be better. If you have the resources, try different methods to determine which works best for your problem.

\item \textbf{Optimize initial configurations.} Our results show that search methods work with different starting setups, but starting with medium-sized networks (e.g., 1024-2048 neurons per layer, 2-3 layers) is often best. Very large networks require longer search times and do not always yield better results.

\item \textbf{Validate on representative data.} The accuracy during the search does not always predict the final test accuracy. The brief duration of the search phase does not adequately showcase the model's ability to generalize. Therefore, it is crucial to evaluate the final model using distinct datasets.

\item \textbf{Leverage efficiency metrics.} When comparing models, consider the accuracy-to-parameter ratio, not just the accuracy. A model with 98.5\% accuracy and 0.5M parameters is often better than one with 98.6\% accuracy and 5M parameters, particularly for devices with limited resources.

\subsection{Summary}
This section presents a comprehensive evaluation of three relaxed bilevel formulations applied to MNIST classification using MLPs and CNNs, as well as CIFAR-10 classification using CNNs. The key findings are summarized as follows.
\end{enumerate}

\begin{enumerate}[label=(\alph*)]
\item All three search methods (NAS-NG, NAS-MA, and NAS-NGMA) successfully identified architectures that maintained competitive accuracy while reducing the number of parameters in MLPs by 40\%--69\% relative to the baseline dense networks, and in CNNs by 50.5\%--75.4\% relative to their respective baselines for MNIST dataset.

\item NAS-NG demonstrated the best practical efficiency, achieving an average parameter reduction of 68.7\% in under one hour of search time while maintaining a test accuracy of 98.53\%. For CNNs, the NG-based model outperformed its baseline by 0.01\% while being approximately 47\% less complex in terms of model size.

\item For MLPs, the NAS-NGMA method achieved the highest accuracy (98.68\%), approaching the best baseline performance while using 75\% fewer parameters, albeit at the cost of substantially longer search times. For CNNs, NAS-NG identified the best-performing model, achieving a test accuracy of 99.63\%.

\item No strong correlation exists between search time and final accuracy, suggesting that computationally expensive search procedures do not consistently outperform simple heuristics for this task.

\item For MLPs, the most efficient discovered architecture achieved 98.49\% accuracy with only 0.45M parameters, corresponding to an efficiency of 218.88 accuracy points per million parameters---approximately 68 times higher than the best baseline. For CNNs, NAS-NGMA emerged as the most efficient approach, achieving approximately 4.06 times the efficiency of the corresponding baseline.

\item Network depth and width interact with search procedures in complex ways, with intermediate depths (3 layers) sometimes outperforming maximal depth, suggesting that search methods can compensate for suboptimal architectural choices.

\item For CIFAR-10, all architectures discovered by the proposed methods outperformed vanilla DARTS, while having significantly smaller model complexity as shown in Table~\ref{tab:overall_performance_cnn_cifar10}.
\item Additional experiments demonstrate that NAS-NG scales effectively to substantially larger architectures, identifying models with higher test accuracy while dramatically reducing parameter counts, and can further improve literature-optimal DARTS architectures through neuron-level optimization, indicating that the proposed strategy complements existing topology search methods.
\end{enumerate}
These findings demonstrate that NAS provides a practical pathway for developing more efficient models without sacrificing performance. The substantial parameter reductions achieved suggest significant opportunities for deploying neural networks in resource-constrained environments while maintaining a strong classification accuracy.

\section{Conclusions}\label{sec6}
This work introduced three relaxed bilevel formulations for gradient-based architecture search---NAS-NG, NAS-MA, and NAS-NGMA---that enable differentiable optimization over neuron- and activation-level design choices in MLPs and CNNs. Across MNIST and CIFAR-10, the proposed methods consistently identify architectures that match or exceed literature baselines while using substantially fewer parameters, confirming that continuous relaxations offer an effective and scalable alternative to discrete combinatorial search. The magnitude of compression achieved, even in simple fully-connected networks, further reveals a level of architectural redundancy that is larger than commonly assumed. Among the three variants, NAS-NG is the recommended starting point for practical use: its sub-hour search times and aggressive compression make it particularly well suited to resource-constrained deployment settings, though discovered architectures should always be validated on held-out data, with evaluation prioritizing accuracy-to-parameter trade-offs over marginal accuracy gains. Additional experiments show that NAS-NG remains effective when initialized from substantially over-parameterized networks and can further refine architectures already optimized by DARTS, indicating that neuron-level optimization is complementary to, rather than a substitute for, topology search.

This study has certain limitations that should be acknowledged. First, results are shaped in part by the simplicity of MNIST, where most architectures reach near-ceiling accuracy (98.40--98.69\% for MLPs), which may obscure differences that would be more pronounced on harder datasets; while CIFAR-10 experiments partially address this, evaluation on larger-scale benchmarks such as ImageNet remains an important next step. Second, the study focuses on depth and width variation in MLPs, leaving interactions with components such as dropout, batch normalization, and skip connections unexplored. Third, hyperparameters were held fixed across methods rather than independently tuned, limiting the precision of cross-method comparisons. Finally, models were retrained from scratch, so potential gains from weight inheritance or knowledge distillation from search-phase models were not examined.

Future work can extend these formulations along several directions: applying them to more challenging vision tasks and a broader range of architectural families (including RNNs and Transformers); incorporating adaptive hyperparameter tuning to improve robustness; and developing a deeper theoretical understanding of why simple gating metrics are effective, which could inform improved heuristics and clarify the problem classes for which such relaxations are best suited. Combining the proposed methods with data augmentation strategies (e.g., AutoAugment) and stronger regularization and optimization techniques may also help close the remaining gap to state-of-the-art accuracy without sacrificing efficiency.

Overall, these findings challenge the assumption that heavy overparameterization is necessary for strong neural network performance, echoing lottery-ticket-style observations and suggesting that targeted, task-specific architectural adjustments can rival more elaborate NAS pipelines. We advocate for a holistic evaluation framework that weighs accuracy jointly with model size and deployment constraints, favoring efficient architectures that deliver practical value over designs optimized solely for marginal accuracy gains.

\section*{Acknowledgments}
The authors gratefully acknowledge the High Performance Computing (HPC) facility at the Indian Institute of Technology (IIT) Kanpur (PARAM Sanganak) for providing the resources that enabled the smooth execution of the experiments in this work.

\bibliography{main}

\appendix
\renewcommand{\thesubsection}{\Alph{subsection}}
\renewcommand{\thesubsubsection}{\Alph{subsection}.\arabic{subsubsection}}

\section*{\centering \Huge Appendix}

\section{Loss Function} \label{loss_fn}
The pre-activation vector of the hidden layer is given by
\begin{equation}
\mathbf{z}_i^{(1)} = \mathbf{W}^{(1)} \mathbf{x}_i + \mathbf{b}^{(1)},
\end{equation}
and the corresponding hidden-layer activations are, $\mathbf{h}_i^{(1)} = \mathrm{ReLU}\!\left( \mathbf{z}_i^{(1)} \right)$. More specifically,
\begin{equation}
h^{(1)}_{ki}
=
\max\!\left(
0,\,
w^{(1)}_{k1} x_{1i} + w^{(1)}_{k2} x_{2i} + b^{(1)}_{k}
\right),
\quad k \in \mathcal{K}_1 = \{1, 2, 3\}
\end{equation}
The pre-activation of the output neuron is
\begin{equation}
z_i^{(2)} = \sum_{k=1}^{3} w^{(2)}_{k} h^{(1)}_{ki} + b^{(2)},
\end{equation}
and the predicted output is
\begin{equation}
\hat{y}_i = \sigma\!\left(z_i^{(2)}\right)
= \frac{1}{1 + \exp\!\left(-z_i^{(2)}\right)}
\end{equation}
Substituting the expressions for $z_i^{(2)}$ and $\mathbf{h}_i^{(1)}$, the predicted output can be written as the following single expression
\begin{equation}
\hat{y}_i
=
\frac{1}{
1 + \exp\!\Bigg(
-\Big[
\sum_{k=1}^{3}
w^{(2)}_{k}\,
\max\!\left(
0,\,
w^{(1)}_{k1} x_{1i} + w^{(1)}_{k2} x_{2i} + b^{(1)}_{k}
\right)
+ b^{(2)}
\Big]
\Bigg)
}
\end{equation}\\
\noindent \textbf{Binary Cross-Entropy Loss.}\\
Given the true label $y_i \in \{0,1\}$ and the predicted probability $\hat{y}_i \in (0,1)$, the loss for the $i^{\text{th}}$ sample is
\begin{equation}
\mathcal{L}_i
=
- \left[
y_i \log(\hat{y}_i)
+
(1 - y_i)\log(1 - \hat{y}_i)
\right]
\end{equation}
Substituting the explicit expression of $\hat{y}_i$, the loss can be written as
\begin{equation}
\mathcal{L}_i
=
- \Bigg[
y_i \log\!\left(
\frac{1}{
1 + \exp\!\big(
- z_i^{(2)}
\big)
}
\right)
+
(1 - y_i)\log\!\left(
1 -
\frac{1}{
1 + \exp\!\big(
- z_i^{(2)}
\big)
}
\right)
\Bigg]
\end{equation}
Let the collection of all network weights and biases be denoted by $\mathbf{W}
=
\{\mathbf{W}^{(1)}, \mathbf{b}^{(1)}, \mathbf{W}^{(2)}, b^{(2)}\}
\in \mathbb{R}^{13}$. We can rewrite the loss as follows
\begin{equation}
\begin{aligned}
\mathcal{L}_i(\mathbf{W}; \mathbf{x}_i, y_i)
=
- \Bigg[
& y_i
\log\!\left(
\frac{1}{
1 + \exp\!\left(
-\left[
\sum_{k=1}^{3}
w^{(2)}_{k}
\max\!\left(
0,\,
w^{(1)}_{k1} x_{1i}
+
w^{(1)}_{k2} x_{2i}
+
b^{(1)}_{k}
\right)
+ b^{(2)}
\right]
\right)
}
\right)
\\[0.5em]
& \hspace{-2.5cm} 
+ (1 - y_i)
\log\!\left(
1 -
\frac{1}{
1 + \exp\!\left(
-\left[
\sum_{k=1}^{3}
w^{(2)}_{k}
\max\!\left(
0,\,
w^{(1)}_{k1} x_{1i}
+
w^{(1)}_{k2} x_{2i}
+
b^{(1)}_{k}
\right)
+ b^{(2)}
\right]
\right)
}
\right)
\Bigg]
\end{aligned}
\end{equation}

\section{Modified Loss Function for Architecture Search} \label{mod_loss_fn}
The sigmoid-based gating is applied after the ReLU activation, and the hidden-layer activations become
\begin{equation}
h'^{(1)}_{ki}
=
\sigma\!\left(\alpha^{(1)}_{k}\right)
\max\!\left(
0,\,
w^{(1)}_{k1} x_{1i}
+
w^{(1)}_{k2} x_{2i}
+
b^{(1)}_{k}
\right),
\quad k = 1,2,3
\end{equation}
Accordingly, the output pre-activation is
\begin{equation}
z_i'^{(2)}
=
\sum_{k=1}^{3}
w^{(2)}_{k}
\sigma\!\left(\alpha^{(1)}_{k}\right)
\max\!\left(
0,\,
w^{(1)}_{k1} x_{1i}
+
w^{(1)}_{k2} x_{2i}
+
b^{(1)}_{k}
\right)
+ b^{(2)}
\end{equation}
Predicted output becomes
\begin{equation}
\hat{y}_i' = \sigma\!\left(z_i'^{(2)}\right)
= \frac{1}{1 + \exp\!\left(-z_i'^{(2)}\right)}
\end{equation}
Given the ground-truth label $y_i \in \{0,1\}$, the modified binary cross-entropy loss for the $i^{\text{th}}$ sample is defined as
\begin{equation}
\mathcal{L}_i(\mathbf{A}, \mathbf{W}; \mathbf{x}_i, y_i)
=
- \left[
y_i \log(\hat{y}_i')
+
(1 - y_i)\log(1 - \hat{y}_i')
\right]
\end{equation}
\begin{equation}
\begin{aligned}
\mathcal{L}_i(\mathbf{A}, \mathbf{W}; \mathbf{x}_i, y_i)= - \Bigg[
&y_i
\log\!\left(
\frac{1}{
1 + \exp\!\left(
-\left[
\sum_{k=1}^{3}
w^{(2)}_{k}
\sigma\!\left(\alpha^{(1)}_{k}\right)
\max\!\left(
0,\,
w^{(1)}_{k1} x_{1i}
+
w^{(1)}_{k2} x_{2i}
+
b^{(1)}_{k}
\right)
+ b^{(2)}
\right]
\right)
}
\right)
\\[0.6em]
& \hspace{-2.5cm}
+ (1 - y_i)
\log\!\left(
1 -
\frac{1}{
1 + \exp\!\left(
-\left[
\sum_{k=1}^{3}
w^{(2)}_{k}
\sigma\!\left(\alpha^{(1)}_{k}\right)
\max\!\left(
0,\,
w^{(1)}_{k1} x_{1i}
+
w^{(1)}_{k2} x_{2i}
+
b^{(1)}_{k}
\right)
+ b^{(2)}
\right]
\right)
}
\right)
\Bigg]
\end{aligned}
\end{equation}
\begin{equation}
\begin{aligned}
\mathcal{L}_i(\mathbf{A}, \mathbf{W}; \mathbf{x}_i, y_i)=- \Bigg[
& y_i
\log\!\left(
\frac{1}{
1 + \exp\!\left(
-\left[
\sum_{k=1}^{3}
w^{(2)}_{k}
\frac{1}{1 + \exp\!\left(-\alpha^{(1)}_{k}\right)}
\max\!\left(
0,\,
w^{(1)}_{k1} x_{1i}
+
w^{(1)}_{k2} x_{2i}
+
b^{(1)}_{k}
\right)
+ b^{(2)}
\right]
\right)
}
\right)
\\[0.6em]
& \hspace{-2.5cm}
+ (1 - y_i)
\log\!\left(
1 -
\frac{1}{
1 + \exp\!\left(
-\left[
\sum_{k=1}^{3}
w^{(2)}_{k}
\frac{1}{1 + \exp\!\left(-\alpha^{(1)}_{k}\right)}
\max\!\left(
0,\,
w^{(1)}_{k1} x_{1i}
+
w^{(1)}_{k2} x_{2i}
+
b^{(1)}_{k}
\right)
+ b^{(2)}
\right]
\right)
}
\right)
\Bigg]
\end{aligned}
\end{equation}

\section{Activation Functions} \label{act_fns}
The activation functions used in this work are summarized in Table~\ref{tab:activation_functions}.
\begin{table}[ht]
\caption{Candidate activation functions are used for neuron activation-based architecture search.}
\label{tab:activation_functions}
\footnotesize
\renewcommand{\arraystretch}{1}
\resizebox{\linewidth}{!}{
\begin{tabular}{@{}>{\raggedright\arraybackslash}p{3.5cm} >{\raggedright\arraybackslash}p{9.0cm}@{}}
\toprule
\textbf{Activation Function} & \textbf{Definition and Mathematical Formula} \\
\midrule

None (Inactive Neuron)
& $\displaystyle f_1(z) = 0$ \newline
Produces a zero output for any input, effectively deactivating the neuron to enable soft pruning within the activation search framework \\

\addlinespace

Identity
& $\displaystyle f_2(z) = z$ \newline
A linear activation that preserves the pre-activation value, allowing the network to retain linear transformations when nonlinearity is unnecessary \\

\addlinespace

Rectified Linear Unit (ReLU)
& $\displaystyle f_3(z) = \max(0, z)$ \newline
Introduces nonlinearity without saturation for positive inputs, enabling effective gradient propagation in deep networks \\

\addlinespace

Leaky Rectified Linear Unit (Leaky ReLU)
& $\displaystyle f_4(z) = \max(0.01z, z)$ \newline
A ReLU variant that retains a small gradient for negative inputs to mitigate the dead-neuron problem \\

\addlinespace

Hyperbolic Tangent
& $\displaystyle f_5(z) = \tanh(z) = \frac{e^{z} - e^{-z}}{e^{z} + e^{-z}}$ \newline
A smooth, zero-centered nonlinear activation mapping inputs to $(-1,1)$ to enhance optimization stability in shallow and medium-depth networks \\

\addlinespace

Sigmoid
& $\displaystyle f_6(z) = \sigma(z) = \frac{1}{1 + \exp(-z)}$ \newline
A bounded, smooth activation mapping inputs to $(0, 1)$, widely used for probability estimation and gating mechanisms \\

\addlinespace

Sigmoid Linear Unit (SiLU)
& $\displaystyle f_7(z) = z\,\sigma(z)= z\ \Bigg(\frac{1}{1 + \exp(-z)}\Bigg)$ \newline
A smooth, non-monotonic activation combining linear and nonlinear behavior to enhance representational capacity in modern networks \\

\bottomrule
\end{tabular}
}
\end{table}

\section{Configuration-Specific Performance Patterns} \label{config_specific}
\begin{figure}[ht]
\centering
\includegraphics[width=0.975\textwidth]{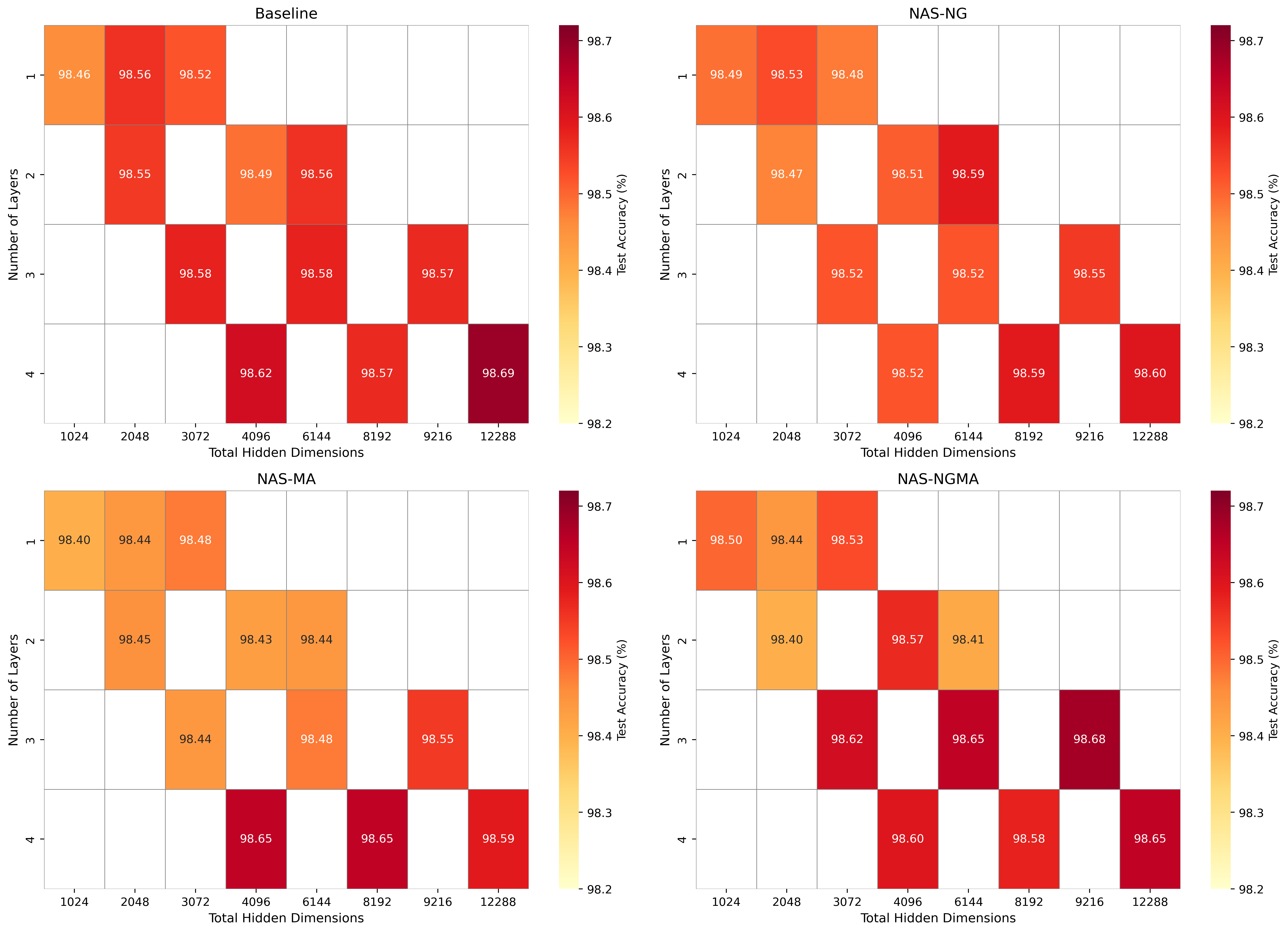}
\caption{Test accuracy heatmaps showing performance on MNIST dataset across different network configurations.}
\label{fig:heatmaps}
\end{figure}

Figure~\ref{fig:heatmaps} presents heatmaps showing test accuracy across the grid of layer counts and total hidden dimensions for each method. Each cell represents the mean accuracy for a specific combination of layer count and total hidden dimensions. Warmer colors indicate higher accuracy. All methods achieve similar accuracy patterns, but searched architectures deliver this performance with fewer parameters.

The heatmaps reveal several interesting patterns. First, all methods show similar overall accuracy distributions, with the highest accuracies generally appearing in deeper networks with larger hidden dimensions (the bottom-right regions of each heatmap). This consistency validates that the search methods are not fundamentally altering the architecture-performance landscape but rather identifying efficient points within it.

Second, the search methods exhibit slightly more uniform performance across configurations. The baseline heatmap shows clearer differentiation between configurations, with noticeable accuracy drops in shallower or narrower networks. In contrast, the search method heatmaps are more uniformly colored, suggesting that architecture search can partially compensate for suboptimal initial configuration choices by identifying the most valuable neurons to retain.

Third, the NGMA heatmap shows the highest peak accuracy (98.68\%) in the 3-layer, high-dimensional region, consistent with earlier findings that this method achieves the best individual results when the initial configuration is favorable.

\section{Additional Analysis} \label{add_analy}
This section presents additional empirical analysis to better understand the behavior of the proposed NAS methods across different architectural configurations. Specifically, we examine the influence of network depth and hidden dimensions on classification performance, along with the parameter reduction achieved relative to the baseline dense architectures. These analyses provide further insights into the trade-off between model complexity and predictive performance.

\subsection{Impact of Network Depth}
\begin{figure}[htpb]
\centering
\includegraphics[width=0.975\textwidth]{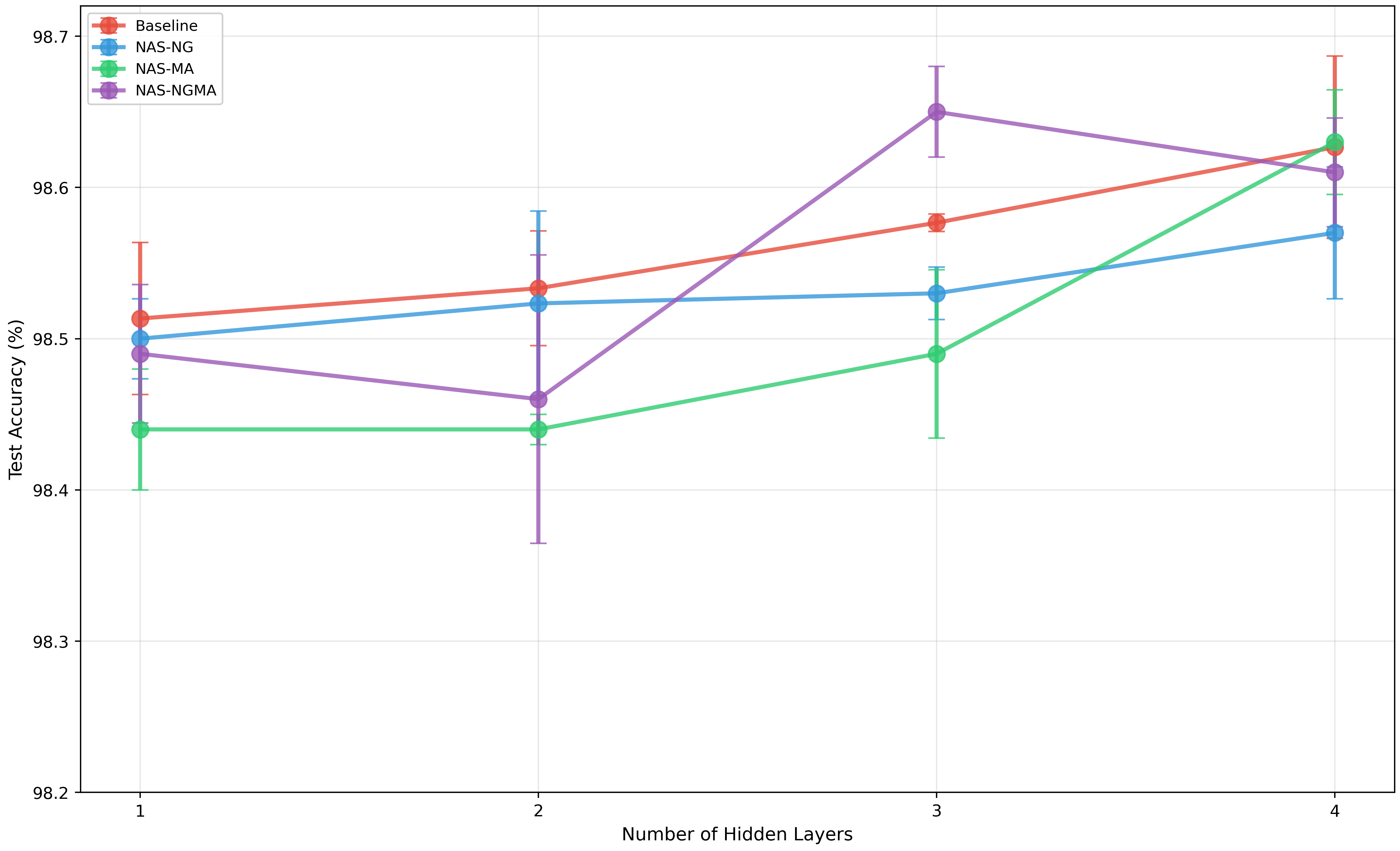}
\caption{Test accuracy vs number of hidden layers, averaged across all hidden dimension configurations.}
\label{fig:acc_layers}
\end{figure}

The relationship between network depth and performance provides insights into how architecture search interacts with fundamental architectural choices. Figure~\ref{fig:acc_layers} illustrates this relationship. Error bars represent standard deviation across the three hidden dimension sizes (1024, 2048, 3072). Deeper networks generally perform better, but the advantage diminishes beyond 3 layers. Search methods maintain competitive performance across all depths.

All methods exhibit a generally positive relationship between depth and accuracy, with 3-layer and 4-layer networks performing best. The baseline shows the strongest depth dependency, improving from 98.51\% (1 layer) to 98.63\% (4 layers). This 0.12 percentage-point improvement demonstrates the value of depth for the MNIST task, though the gains are modest given the dataset's relative simplicity.

Interestingly, the search methods show less dramatic variation across depths. The NG approach maintains remarkably consistent performance (98.50\% to 98.57\%), suggesting that the pruning process adapts well across different architectural configurations. NAS-NGMA method shows the most interesting pattern, achieving its best performance (98.65\%) at 3 layers rather than 4, indicating that the interaction between gating and activation criteria may favor intermediate depths where both metrics provide clear signals.

\subsection{Influence of Hidden Dimensions}
\begin{figure}[htpb]
\centering
\includegraphics[width=0.975\textwidth]{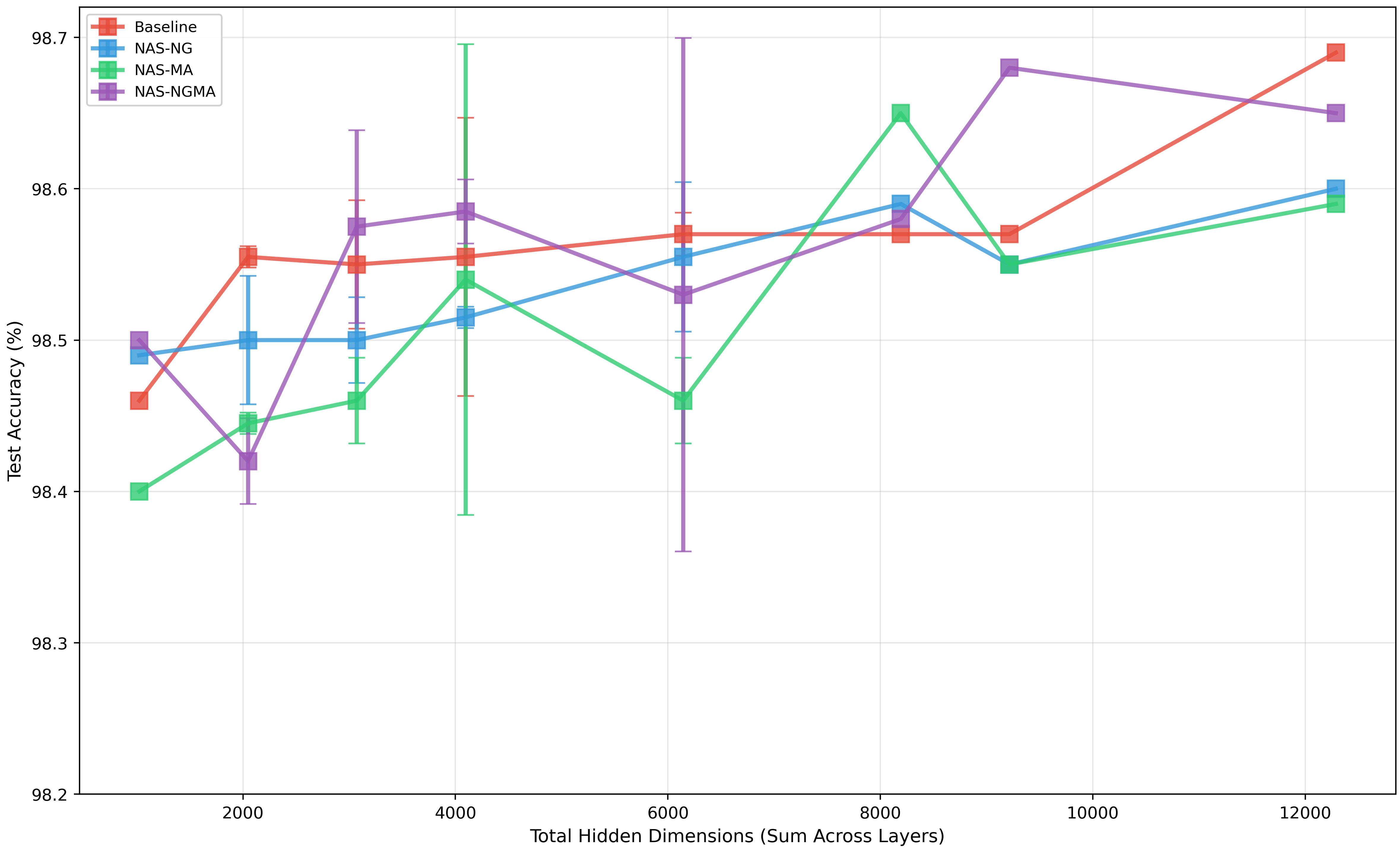}
\caption{Test accuracy versus total hidden dimensions, averaged across configurations with the same total capacity.}
\label{fig:acc_dimensions}
\end{figure}

Figure~\ref{fig:acc_dimensions} examines how the total hidden dimension capacity (sum across all layers) affects performance. The standard deviation across various layer configurations that result in identical total dimensions is represented by error bars. Larger hidden dimensions generally improve accuracy, but with diminishing returns beyond 6,000 total dimensions.

The results reveal diminishing returns as the number of hidden dimensions increases. Moving from 1,024 to 4,096 total dimensions yields noticeable improvements in accuracy across all methods. However, further expansion beyond 6,000 dimensions provides minimal additional benefit, with some configurations even showing slight performance degradation. This pattern is particularly evident in Figure~\ref{fig:acc_dimensions}, where the largest configurations (12,288 total dimensions for 4-layer networks with 3,072 neurons each) do not consistently outperform smaller configurations. This observation validates the architecture search approach: rather than simply scaling up to massive networks, intelligent pruning can identify the essential capacity needed for the task.

The search methods follow similar trends but with reduced overall parameter counts at each dimension level. Notably, NAS-NG achieves competitive accuracy even at lower total dimensions, suggesting that it successfully identifies the most informative neurons rather than simply retaining neurons in proportion to the initial size.

\subsection{Parameter Reduction Analysis}
\begin{figure}[htpb]
\centering
\includegraphics[width=0.975\textwidth]{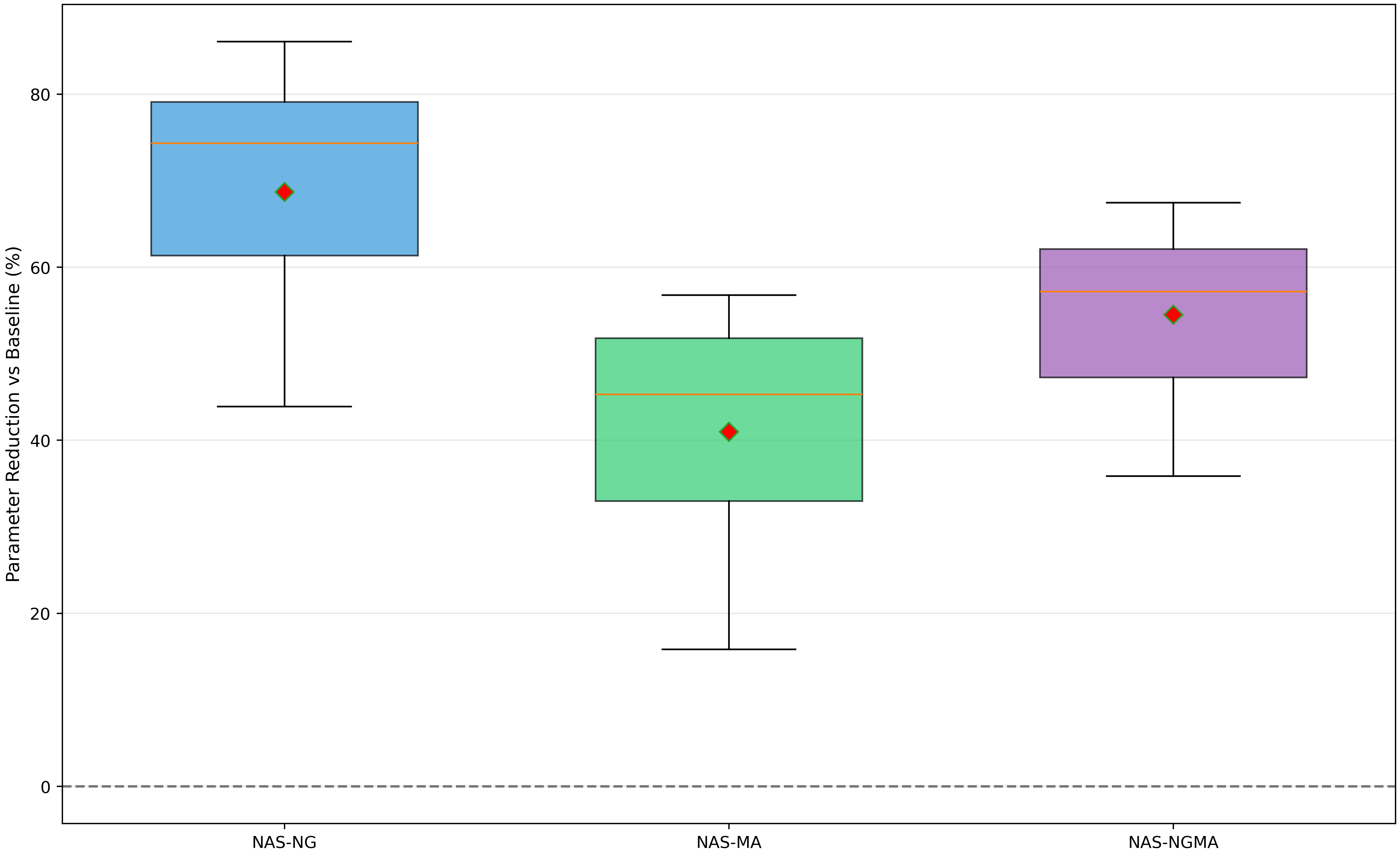}
\caption{Distribution of parameter reduction percentages achieved by each search method compared to baseline dense networks of equivalent initial size.}
\label{fig:param_reduction}
\end{figure}

To quantify the efficiency gains more precisely, we analyzed the parameter reductions achieved by each search method relative to its corresponding baseline configuration. Figure~\ref{fig:param_reduction} presents this analysis in a box plot format. The red diamond markers indicate mean reduction values.
NAS-NG achieves the highest and most consistent parameter reductions, averaging 68.67\% (range: 43.88\%--86.06\%, std.\ dev.\ 15.64\%). This reliable compression across configurations highlights the effectiveness of NG-based NAS.

The combined approach yields a more conservative mean reduction of 54.49\% (std.\ dev.\ 9.77\%), reflecting its multi-criteria decision-making. By requiring agreement between gating and activation metrics, it is less likely to make extreme pruning decisions that might work well for one criterion but poorly for another.

MA-based pruning shows a relatively smaller average reduction of 40.98\% and greater variability (15.82\%--56.76\%, std.\ dev.\ 14.61\%), indicating configuration-dependent performance that works well for some architectures but less effectively for others. A similar reduction in model complexity is also observed for the CNN architectures, as reported in Tables~\ref{tab:overall_performance_cnn} and ~\ref{tab:overall_performance_cnn_cifar10}.
\end{document}